# Formation of structural attractors in neuromorphic systems

Yurii Parzhyn[1], Alexander Schwarzmann[1], Mykyta Lapin[2], Kostiantyn Bokhan[2]

[1] Augusta University, Augusta, USA

[2] National Technical University "Kharkiv Polytechnic Institute", Kharkiv, Ukraine

## Abstract

This paper examines the theory of Invariant Structural Learning (ISL), which proposes a non-optimization approach to concept formation. Learning is interpreted as convergence to structural attractors in a hypergraph space, rather than as the minimization of a global loss function. The paper presents the ISL model, including its mathematical formalization, computational verification, and a hypothetical neurobiological interpretation. The mathematical section introduces the formal apparatus of the structural reduction process and proves its finite convergence, the existence and uniqueness of class structural attractors, and the self-organization of attractor maps. The computational section demonstrates the feasibility of the proposed approach on classical image recognition tasks, utilizing the proposed learning mechanism without backpropagation and with extremely small training datasets. Finally, the neurobiological section formulates hypotheses regarding the possible implementation of structural attractors in dendritic trees, neural coding as a projection of internal attractor dynamics, and the development of neural architectures supporting the proposed learning concept. These hypotheses are discussed in the context of modern experimental data in the fields of dendritic computations, synaptic plasticity, and the structural organization of neural circuits. The proposed neurobiological mechanisms are presented as testable hypotheses rather than established biological facts. The results demonstrate the mathematical consistency and computational feasibility of the proposed model, while the neurobiological hypotheses outline potential directions for its experimental verification.

## I. Introduction

### 1. Relevance and Objectives of the Study.

The modern machine learning paradigm relies almost entirely on the optimization of continuous error functions using the backpropagation algorithm (BackProp) or its modifications. Statistical approaches to learning, in the sense of Vapnik-Chervonenkis theory and Leslie Valiant's PAC-learning (Probably Approximately Correct), assume the existence of an external quality criterion, and the process itself is formalized as the minimization of the discrepancy between a prediction and a given objective function. Despite the outstanding practical successes of deep neural networks, this paradigm has fundamental limitations: dependence on the choice of loss function, sensitivity to data distribution, the need for significant training sets and learning epochs, lack of transparency in the decision-making process, and high computational and other costs. Furthermore, this approach exhibits a deep divergence from the principles of real biological neurons and neural systems: it requires a global error signal and its retrograde distribution across hidden neurons (*the credit assignment problem*), symmetry of forward and backward connections (*the weight transport problem*), and the division of processes into forward and backward passes, which is impossible in continuously functioning living tissue. Moreover, classical artificial neural networks (ANNs) abstract the neuron to the level of a simple summation point for weighted inputs, completely ignoring the complex, functionally coupled spatial structure of real cells.

Experimental data in neuroscience prove that the brain functions differently. The dendritic tree of a single pyramidal neuron is not merely a summer, but a powerful nonlinear "computational device" capable of locally processing complex spatiotemporal patterns of synaptic activity. Learning in living systems is endogenous in nature. At the level of individual neurons, it is governed not by minimizing a global external loss function, but by local plasticity mechanisms (such as synaptic tagging, capture, etc.), as well as large-scale synaptic pruning - the directed reduction of redundant connections established during morphogenesis. The brain does not "fit weights" to an external objective function - it carries out continuous structural self-organization, forming stable invariant representations even on the basis of very limited experience (Few-Shot Learning).

We are convinced that true understanding in intelligent systems is not a statistical mapping of inputs to outputs, but a process of interpreting input information through an internal "world model" formed by the system itself. This conviction inspires us to develop an alternative theory of learning for biologically motivated ANNs: *Invariant Structural Learning* (ISL), which formalizes known biological principles at the intersection of hypergraph algebra and dynamical systems theory. We assume that learning in biological neurons can be viewed as a directed reduction of a physical hypergraph implemented in the structure of the dendritic tree. Knowledge in such a system is represented not by a distribution of parameters or a probabilistic estimate, but by a stable invariant structure arising from self-organization.

This stable invariant structure is the key concept of the proposed theory - *the structural attractor*. From the perspective of psychology and cognitive linguistics, a structural attractor is the endogenous equivalent of a "*Concept*" in its ontological sense. We hypothesize that in the brain, this structure is implemented by the dendritic tree of a pyramidal neuron. Unlike classical attractors in dynamical systems, this object is defined not by trajectories in a continuous space, but by the formation of structural-parametric invariants during the reduction process.

In our view, the proposed approach radically changes the computational basis of ANNs:

- *Complete abandonment of BackProp and gradient descent*: Learning is implemented as a monotonic dissipative process of structural-parametric reduction, governed by local rules;
- *Biologically adequate data efficiency*: Stable attractor-concepts are formed on extremely small datasets without preliminary or meta-learning, or the application of heuristic rules, and exclusively using positive examples in a single pass (one epoch);
- *Local explainability (Explainable AI)*: The topology of each attractor is transparent, which allows for a mathematically rigorous interpretation not only of the classifier's correct decisions but also of the reasons for its errors.

Thus, we propose a new, non-connectionist approach to constructing and training biologically motivated – neuromorphic models of neurons and neural networks. By "*neuromorphic,*" we mean a property of the modeled system in which information processing principles - regardless of the physical representation of the signal in the response of the simulated neuron - are structurally isomorphic to the organization of biological neural tissue at the following levels:

– *Co-located memory-computation*: The synapse simultaneously stores state (weight) and participates in computation, unlike the separated memory and ALU in the von Neumann architecture;
– *Local adaptivity*: Changes in synaptic efficacy depend on locally accessible signals of pre- and postsynaptic activity, rather than on a global error signal calculated externally;
– *Massive parallelism and distributed representations*: Computation is realized not by a sequence of instructions, but by the simultaneous activity of a multitude of elements;
– *Robustness to noise and degradation*: Achieved through redundancy rather than through precise arithmetic calculations as in connectionist ANNs.

The objective of this work is a formal presentation of the mathematical apparatus of ISL, a proof of the existence of the structural attractor based on theorems regarding reduction convergence and the uniqueness of the class attractor, an experimental demonstration of the model's potential on classical pattern recognition tasks (Proof-of-Concept (PoC)), and a hypothetical neurobiological interpretation of the proposed theory.

The proposed theory does not claim to provide a direct description of the biological mechanisms of brain function. However, it relies on a number of well-known neurobiological observations related to the role of dendritic computations, the selection of stable structures, and the formation of long-term representations. In this regard, the model can be viewed as a biologically motivated mathematical abstraction intended for investigating the general principles underlying the formation of internal models of the "external world."

**2. Research Background.**

This work is a continuation and deepening of the theoretical and experimental results presented in [1 - 4]. It draws upon the fundamental works of David Hubel and Torsten Wiesel (D. Hubel "*Eye, Brain, and Vision"*, 1988), David Marr ("*Vision"*, 1982), and John Hopfield ("*Neural networks and physical systems with emergent collective computational abilities*," 1982), as well as a long-standing tradition of structural pattern recognition that achieved significant theoretical and practical milestones prior to the widespread adoption of convolutional neural networks (CNNs).

For instance, Xiao et al. (2008) [5] established that *the topological structure of an image can serve as a class invariant*. The authors constructed a hierarchical description of images that is invariant to their visual style. Their technological pipeline consists of the following stages:

1. *Primitive Extraction:* DoG (Difference of Gaussian) regions are used and filtered via a greedy algorithm based on size to ensure consistent coverage of both line drawings and photographic volumes.
2. *Tree Construction (Hierarchy):* Neighboring regions are iteratively merged into a tree structure based on their statistical similarity (Agglomerative Merging).
3. *Segmentation via Graph "Energy":* To split the tree into an "object" and "its parts," the authors use the Gutman "Laplacian Graph Energy" functional, seeking local minima. The graph with minimal "energy" corresponds to the most regular, natural structure of parts (e.g., distinguishing the head, body, and legs of a horse in an image).
4. *Feature Vector Formation:* Part topology is encoded as a fixed vector via spectral graph analysis (eigenvalues of the Signless Laplacian matrix). Thus, the graph structure is ultimately compressed into a flat, fixed-dimensional (8D) vector, which is then fed into a statistical classifier (Gaussian Mixture Model, GMM). Such a mapping inevitably loses information regarding the relative positioning of graph elements, as different structures may share an identical or similar Laplacian spectrum.

5. *Classification:* The resulting structural vectors are clustered using a standard GMM.

The authors conducted an experiment on an image dataset of 12 classes (faces, birds, flowers, horses, cows, etc.) containing significant stylistic diversity. The model was trained in a real Few-Shot mode - using only 10 examples per class. The results showed high accuracy: for example, 94% for faces and animals. However, it should be noted that 10 points in an 8D space represent an extremely sparse sample: a Gaussian constructed from 10 points will be either excessively broad (letting in false objects) or excessively narrow (causing overfitting). Thus, the authors build statistics not around class variability, but around the density distribution of 8D vectors. With only 10 vectors, the GMM attempts to describe them as a compact, continuous "cloud" (a Gaussian) in 8D space.

Despite remaining within the statistical learning paradigm, Xiao et al. demonstrated that if one isolates a stable structural organization of an image through hierarchical aggregation (in their case, DoG regions and weighted Laplacian matrices), one can obtain a clean, separable topological structure as the output.

Another work by Geurts et al. (2006) [6] proposes a generic mathematical framework for training invariant classifiers on any topologically structured data whose elements are connected by a neighborhood relation: pixels in images, letters in strings, or samples in time series. Their algorithm consists of three deterministic steps:

1. *Splitting:* A complex object (image, DNA, signal) is cut into local "pieces" of a specified diameter.
2. *Learning:* A base algorithm is trained to predict the object’s class based on its individual pieces.
3. *Aggregation:* The final decision for a new object is made through "voting" or averaging the predictions of all its pieces.

At the time, Geurts et al. demonstrated State-of-the-Art (SOTA) results in three distinct fields:

- *Bioinformatics (Strings):* Classification of DNA sequences (Splice and MSN2 datasets). The method identified hidden local patterns without biological heuristics.
- *Gesture Recognition (Time Series):* The Auslan-s dataset (sensory glove data for Australian Sign Language). The model showed an error rate of only 1.0%, surpassing the existing SOTA record (1.5%).
- *Computer Vision (Images):* Testing on the extensive medical IRMA database (10,000 images, 57 anatomical zone classes). The random window method showed high accuracy while automatically highlighting areas for doctors that influenced the decision.

In this work, Geurts and his co-authors made the same compromise as Xiao:

1. *Statistical Bottleneck:* After "cutting" the object into pieces, the authors feed them into an ensemble of Extra-Trees and use statistical probability averaging (soft voting) at the end, reverting to statistical ML.
2. *Lack of Connection Hierarchy:* The proposed method is permutation-invariant. It simply dumps all parts into a "bag" and counts votes, completely losing information about how these pieces are connected to one another—i.e., the algorithm does not "see" the scene's geometry. Thus, having proven the universality of topological data parts, the authors package them into the statistical structure of an ensemble.

An analysis of these works shows that the authors empirically demonstrated that structure and local topological invariants can be the key to building a "world model." However, in both approaches, the structural description of the object is used only as an intermediate representation, after which the recognition task is reduced back to a statistical estimation of feature distributions in Euclidean space.

Thus, structure serves as a source of features but does not become an independent object of learning, whereas in our proposed ISL theory, the structural attractor is itself the result of learning and the criterion for recognition. In our work, we shift the focus from the construction of structural features to explaining the mechanisms of their emergence.

It is also necessary to note that there are many models attempting to overcome the barrier of biological inadequacy in the classical statistical approach to learning. Consider one of the most recent biologically motivated models: Dendritic Localized Learning (DLL) by Changze Lv, Jingwen Xu, et al. [7]. The authors claim to have overcome three main limitations:

- the need for weight symmetry during forward and backward signal propagation;
- the use of global error signals;
- the two-phase nature of learning.

Indeed, the DLL model introduces a matrix $\Theta$ (backward weights), which is not a transposed copy of $W$ (forward weights) but rather another set of trainable parameters. However, the matrix $\Theta$ is not dynamically independent. Although it does not coincide with the transposed matrix $W^T$, its update is determined by the same local errors as the update of the forward weights (formulas (7) - (9) in their paper). Consequently, matrices $W$ and $\Theta$ form two interconnected subsystems of a single learning algorithm rather than two independent parameter systems. Thus, the authors eliminated the geometric constraint (symmetry) but failed to eliminate the functional dependency between forward and backward connections.

Furthermore, the local error $\xi_i$ proposed in the work is not an independent quantity. According to formula (7) in their paper, it is recursively calculated from the error of the subsequent layer $\xi_{i+1}$, which in turn is determined by the error of an even higher layer. Thus, the entire chain of local errors is generated by a single error at the output layer arising from comparing the network output with the target label. Therefore, the source of error information (the global error) is not eliminated but merely transformed into a sequence of local errors (expected value). This preserves the recursive *credit assignment principle* underlying the backpropagation family of algorithms. In other words, in DLL, *the credit assignment problem* has not disappeared; the authors simply propose a different mechanism for its implementation.

Regarding the elimination of two-phase learning: the authors claim that forward signal propagation and learning can occur simultaneously. However, formula (4) in their work indicates that even if calculations overlap in time, their logical dependency remains; i.e., it is not the two-phase nature (as a dependency of calculations) that disappears, but merely their strict temporal organization.

Thus, DLL eliminates the explicit temporal separation of forward and backward passes, yet the informational dependency between them persists. The calculation of each layer's local error is possible only after the error of the subsequent layer is formed, so the causal sequence of error information propagation remains unchanged. Despite the claimed biological motivation, the main contribution of DLL lies not in abandoning the error-driven learning paradigm but in modifying the mechanism of its propagation. The algorithm eliminates the weight symmetry requirement but retains recursive error information propagation and the gradient principle of parameter adjustment. Consequently, DLL represents an evolutionary development of the classical statistical learning paradigm rather than a transition to a fundamentally different learning model.

In our proposed model, learning is formulated without the use of error signals or the formation of predictions characteristic of the error-driven learning family. Instead, the learning process is viewed as the sequential structural self-organization of the system's internal representation.

## II. Mathematical Foundations of the Model

### 1. Base Spaces and Structures.

1.1. Input Space $\mathcal{X}$**.** This is the input data space containing "raw" objects. Examples include MNIST images, geometric shapes, object contours, and 2D or 3D models.

1.2. Space of Atomic Structural Elements (Primitives) $\mathcal{E}$**.** Examples for MNIST: line segments (assuming the contour is approximated by segments); bifurcation points of the contour structure: endpoints, corner points, and points of intersection or conjugation of segments. These are the "bricks" from which any representation of input patterns is constructed.

1.3. Space of Hypergraphs (Representations of a Single Model Neuron)**.**

$$\mathcal{Z} = \{\mathcal{G} = (E, \mathcal{R}^{(s)}, \mathcal{R}^{(p)})\} \quad (1),$$

where:

– each element $z \in \mathcal{Z}$ is a hypergraph $\mathcal{G}$ describing a holistic class image;

– $E \subseteq \mathcal{E}$ is a finite set of structural elements (vertices of the graph/hypergraph);

– $\mathcal{R}^{(s)}$ represents spatial edges (edges of the base spatial/temporal graph) that define the connectivity of the structure;

– $\mathcal{R}^{(p)}$ represents parametric hyperedges/hyperarcs that define the interrelationships between the parameters of different structural elements of the hypergraph. The hypergraph is formed as a "superstructure" over the base spatial graph. Each vertex of the base graph is interconnected with its own set of parameters. In this way, a *stratification (fibration) of the system's parametric space* occurs.

## 2. Extraction and Measurements

2.1. Primitive Detector.

$$D: \mathcal{X} \to \mathcal{Z}_0, \mathcal{G}_0 = D(x) \quad (2),$$

where:

– $D$ is the primitive detection/extraction operator. This operator models the operation of the sensory system. The detection stage results in the formation of the structural elements of the hypergraph. The construction and study of primitive detectors are not the goal of this work. For simplicity of modeling, we assume they are given.

2.2. Measurements.

$$m_i: \mathcal{E} \to M_i, m(\mathcal{G}) = \{m_i(e) \mid e \in E\}_{i=1}^{k} \quad (3),$$

where:

– $m_i$ is the measurement function of the $i$-th parameter of element $e$;

– $m(\mathcal{G})$ is the set of all measured parameters of the entire graph.

Measurements result in the formation of parameters - modes of the structural elements. Each structural element in the hypergraph is associated with its *own modal group*: *a set of measured parameters*. Parameters and their values form a "superstructure" over the base structural element (e.g., a line segment or a point of a contour image). Thus, a connected multi-dimensional parametric space of the system (model) is formed. It is this parametric space that establishes the structure of the hypergraph representation of the perceived image, where parameters are represented as vertices/hypervertices (in an alternative representation, parameters can define the labeling of hyperedges or hyperarcs). This parametric space is presented in the form of exogenously specified coordinate systems and measurement scales. The partitioning of parametric spaces into generalized segments during the learning process forms the metric of these spaces. Exogeneity, in this case, is the price paid for modeling. Examples of parameters for a skeletonized MNIST contour include: segment length, segment orientation direction, the magnitude of the angle between segments, and point coordinates.

### 3. Reduction

The core premise of the entire proposed theory is the assumption that within any structure of examples belonging to a single class, there exists a hidden or "noisy" structural attractor that determines the stable invariant structure and properties of the entire class. This structural attractor manifests itself through the reduction process.

3.1. Parametric Reduction $R_p$.

$R_p$ is the operator of ordering and compression within metrics. This operator does not change the spatial structure of the graph/hypergraph, but it alters the space of parametric relations.

$$R_p(\mathcal{G}) = \left(E, \mathcal{R}^{(s)}, \mathcal{R}_Q^{(p)}\right) \quad (4),$$

$$\mathcal{R}_Q^{(p)} := Q\big(m(\mathcal{G})\big) \quad (5),$$

where:

- if $\mathcal{R}^{(p)}$ - relations between elements through parameters $m_i(e)$ before reduction, then $\mathcal{R}_Q^{(p)}$ represents the relations between elements via segments $Q\big(m_i(e)\big)$ after reduction. We replace continuous values with discrete classes (segments);
- $Q$ is a clustering (quantization) operator over metric $\rho_i$ with a stability parameter/threshold $f$ (discussed below), which practically sets the boundaries of the coordinate system segments or scales. In other words, this operator partitions the parameter space $M_i$ into stable segments $\Sigma_i$ according to metric $\rho_i$ with stability threshold $f$. Ultimately, parameters can "compress" down to qualitative categories or disappear from the hypergraph structure.

If:

$$m_i(e) \in M_i \quad (6),$$

then:

$$Q_i\big(m_i(e)\big) \in \Sigma_i \quad (7),$$

$$Q(m(\mathcal{G})) = \{Q_i\big(m_i(e)\big)\}_{e,i} \quad (8),$$

$$Q_i = \text{cluster}_{\rho_i,\epsilon}: M_i \to \Sigma_i \quad (9),$$

where $\rho_i$ is the metric in the parameter space $M_i$, which determines the segmentation structure:

$$\rho_i: M_i \times M_i \to \mathbb{R}_{\geq 0} \quad (10).$$

***Axiom 1.*** *Stability (idempotency) of segmentation:*

$$Q\big(Q(m)\big) = Q(m) \quad (11).$$

Segmentation is stable; its re-application changes nothing.

3.2. Structural Reduction $R_s$.

Structural reduction is performed in several stages:

1. *A frequency measure is determined* - how many times a structural element of the hypergraph (its structural or parametric vertex/hypervertex) occurs in all examples of the class:

$$f(e) = \frac{1}{N}\sum_{j=1}^{N} \mathbf{1}[e \in \mu_j(E_j)] \quad (12),$$

where:

– $E$ is the space after alignment;

– $\mu_j$ representing the partial matching of the *anchors* of the $j$-th example;

– **1** is an indicator function (1 - if the condition is true; 0 - otherwise),

– $N$ is the number of accumulated class examples.

2. *The structural selection operator is applied - selecting stable structural elements* $S(E)$*:*

$$S: 2^E \rightarrow 2^E \quad (13),$$

$$S(E) = \{ e \in E \mid f(e) = 1 \} \quad (14),$$

where:

– $\{e \in E \mid f(e) = 1\}$ - elements occurring in absolutely all class examples after alignment (construction and matching of hypergraphs) are *irreducible*;

– $2^E$ - the set of all subsets of the set $E$. The operator $S$ assigns to each set of structural elements its reduced subset, containing only invariant (and anchor) elements.

We define $A$ as the *set of anchors* - hypergraph elements that appear in all examples (for the MNIST example under consideration). This set of anchors is updated dynamically during the training process after each new example (when computing $f(e)$).

If the training set contains examples where the structure of *the hidden attractor* is incomplete (a problem that can occur in real systems with unprepared training sets), it becomes necessary to weaken the condition (12): $f(e) \geq \tau$, where $\tau$ is a threshold value for the frequency of occurrence/stability or truth of structural elements or parameter values. In the model, $\tau$ can be considered as an empirical hyperparameter that may vary, for example, within the range $\tau \approx 0.8 - 1.0$. However, for simplicity of exposition in this paper, we will use a prepared training set where all examples have the complete structure of the hidden attractor.

If this problem of input pattern incompleteness arises during inference, it is resolved using *associative recognition*. *Autoassociative* and *heteroassociative* recognition exist as separate mechanisms, which is confirmed by numerous studies both in neurosciences (David Marr (1971), Kazu Nakazawa et al. (2002), etc.) and computer science (Shun-ichi Amari (1972), Teuvo Kohonen (1972), John Hopfield (1982), Stephen Grossberg & Michael Cohen (1983), etc.). Therefore, considering this issue within ISL is beyond the scope of this work.

In other words, $S(E)$ is a filter that removes all unstable elements and retains only stable, critically important structural elements/parameters - the "*anchors*." For MNIST, these "anchors" include, for example, critical bifurcation points of the image contour (endpoints or corner points), among others. It is $S(E)$ that makes it possible to transform the initial memorized graph into the minimal structure of the attractor.

***Definition 1.*** An element $e \in E$ is called an *anchor* if its frequency measure satisfies the condition $f(e) = 1$ (or $f(e) \geq \tau$):

$$A := \{ e \in E \mid f(e) = 1 \} \quad (15),$$

$$S(E) := A \quad (16).$$

An anchor is a structural element (such as a vertex or a hyperedge of a hypergraph) that is stably present in all (at $f = 1$) or almost all (at $f \geq \tau$) examples of a class. It plays the role of a "reference point" of the structure around which the construction and alignment of the hypergraph structure take place. Anchors are not removed during reduction and are "glued together" by structural elements (edges/hyperedges) to preserve structural integrity after the removal of non-critical vertices (attractor "noise").

*Uniqueness of the set of anchors $A$.* If the anchor invariance criterion for the accumulated configuration of a class of objects $\mathcal{C} \subset X$ is fixed (in this case, it is the frequency measure), then the set of anchors $A$ is *uniquely determined by the accumulated structure*.

3. *Local Phase of Structural Reduction.*

Let:

$$\mathcal{G} = \left(E, \mathcal{R}^{(s)}, \mathcal{R}^{(p)}\right) \quad (17)$$

is the accumulated structural configuration.

We define *the local reduction operator*:

$$L(\mathcal{G}) = \left(A, \mathcal{R}^{(s)}|_A, \mathcal{R}^{(p)}\right) \quad (18).$$

Here, $\mathcal{R}^{(s)}|_A$ denotes the restriction of structural relations to the set of anchors $A$.

The operator $L$ performs two operations:

- removes non-anchor elements $E \setminus A$;
- preserves the structural and spatial relations associated with the remaining anchors.

At the same time, $L$ does not restore broken connectivity.

4. *The process of "gluing" the structure after reduction* is determined by the operator of restoration or closure of the connectivity "Closure" over the remaining (minimal set) structural elements to preserve the connectivity of anchors in a single hypergraph.

Let the set of all anchor elements of a class be denoted as $A \subseteq E$.

***Definition 2.*** *The spatial closure operator, called the $G$ operator* (Closure $= G$), is an operator that restores structural connectivity between the remaining anchors in accordance with the base spatial relations $\mathcal{R}^{(p)}$.

For each valid set of anchors $A$ and a fixed spatial structure of the accumulated configuration, the operator $G$ returns a unique structure on $A$ defined by the spatial closure rules given by this fixed spatial structure. If this structure allows for multiple alternative ways of restoring connectivity (for example, different directions of traversing the image contour), then the corresponding selection rule is part of the definition of $G$.

In other words, if the removal of non-anchor elements breaks the spatial chain between two or more anchors, Closure induces a corresponding structural relation between them that preserves the original spatial sequence. The resulting structure:

$$\mathcal{G}^* = G\left(A, \mathcal{R}^{(s)}|_A, \mathcal{R}^{(p)}\right) \tag{19}$$

contains only anchor elements and the relations between them, restored while preserving the underlying spatial connectivity.

The Closure operator possesses the following properties:

1. *Determinism*.

Let the set of anchors $A$ be fixed, and let the base spatial structure $\mathcal{R}^{(p)}$ be fixed as well. Then the result $G\left(A, \mathcal{R}^{(p)}\right)$ is uniquely determined.

Indeed:

– Closure does not make choices among alternative candidate structures, because it uses only the fixed values of $A$ and $\mathcal{R}^{(p)}$;

– for each pair or sequence of anchors between which the original spatial structure transitively defines a continuous path through the removed non-anchor elements, Closure restores the corresponding relation between the anchors. Consequently, the result of Closure is uniquely determined:

$$G\left(A, \mathcal{R}^{(p)}\right) = \mathcal{G}^* \tag{20}$$

2. *Idempotency.*

Let $\mathcal{G}^* = G\left(A, \mathcal{R}^{(p)}\right)$ is the result of spatial closure.

Then the repeated application of Closure does not change the structure:

$$G(G(\mathcal{G}^*)) = \mathcal{G}^* \quad (21).$$

Indeed, after executing Closure, all relations necessary to preserve spatial connectivity between anchors have already been restored.

The repeated application of Closure does not detect deleted non-anchor elements, since they do not belong to $A$, and it does not create new relations, since the initial spatial connectivity is already closed.

The Closure operator does not introduce new spatial or parametric connections. It merely compresses existing chains of spatial connections into equivalent higher-level connections.

Consequently, formula (21) holds true.

Then, the full structural reduction operator is defined as:

$$R_s = G \circ L \quad (22).$$

Consequently:

$$R_s(\mathcal{G}) = G\left(L(\mathcal{G})\right) \quad (23).$$

The operator $R_s$ sequentially performs:

accumulation → anchor identification → pruning of non-anchor elements → spatial closure.

Thus, local reduction and Closure are two distinct phases of a single structural transformation.

***Lemma 1.*** *Finiteness of structural reduction.*

Let the initial structural configuration contain a finite number of elements.

If each step of local reduction removes at least one non-invariant element or reaches a state where all remaining elements are anchors, then the local reduction process terminates in a finite number of steps.

*Proof.*

Consider the number of remaining non-anchor elements: $B_t = |E_t \setminus A_t|$.

At each step that continues the reduction, $B_{t+1} < B_t$. Since $0 \le B_t \le |E_0|$, a strict decrease in $B_t$ cannot continue infinitely. Consequently, there exists a finite moment $T$ such that $B_T = 0$. At this moment, all remaining elements are anchors, and the local reduction stabilizes. ■

***Definition 3.*** *Structural-canonical functional of anchors.*

Let $E = E_p \cup E_v$ be the partition of the set of structural elements into point ($E_p$) and vector ($E_v$) primitives.

The functional $\varphi$: Z → $2^E$ is called *structural-canonical* if:

$$\varphi(\mathcal{G}) = \{\, e \in E_P \mid \tau(e) \in \mathcal{T}^* \,\} \qquad (24),$$

where:

– $\tau(e)$ is the type of the structural element $e$, determined locally from the topology of the base spatial graph;

– $\mathcal{T}^*$ is the set of structurally significant types under the following conditions:

1. Locality: $\tau(e)$ is determined only by the local structure of the neighborhood of $e$ in the base spatial graph and does not depend on the history of construction $\mathcal{G}$;

2. Isomorphism invariance: if $\mathcal{G}_1 \sim \mathcal{G}_2$, then $\varphi(\mathcal{G}_1) \sim \varphi(\mathcal{G}_2)$;
3. Reduction stability: if $\mathcal{G}' \preceq \mathcal{G}$, then $\varphi(\mathcal{G}') \subseteq \varphi(\mathcal{G})$.

Then we can refine formula (12):

$$f(e) = \frac{1}{|\ \mathcal{C}\ |} \sum_{x \in \mathcal{C}} 1[e \in \varphi(H_x)] \tag{25}$$,

where:

– $\mathcal{C} \subset X$ is the object class;

– $\varphi$ extracts structural elements of types from $\mathcal{T}^*$ based on the local topology of the neighborhood of $e$, independently of the matching procedure, i.e., it maps the local element of each $H_x$ into a common space of types;

– $e$ in this case is not a specific vertex of a specific graph, but a structural equivalence class / canonical type.

Formula (25) preserves the meaning of the frequency measure of an element's presence in class examples, but eliminates the dependence on the matching algorithm (12).

Formula (12) defines the empirical frequency of a structural element on a finite training set $N \subseteq \mathcal{C}$. Formula (25) introduces the corresponding population frequency on the full class $\mathcal{C}$, which is necessary to define the ideal anchor structure $A_{\mathcal{C}}$.

Further, it is shown that under the conditions of structural completeness (completeness of the hidden structural attractor) and sufficient diversity of the training set (structural and parametric variability sufficient to form the attractor as a fixed point of reduction), the reduction dynamics reconstruct this structure from a finite $N$, i.e., $A_{\mathcal{C}} = A_N$. Therefore, $f_{\mathcal{C}}$ is used in the theoretical description of the invariant structure of the class, while $f_N$ is used in the operational learning process.

Hereafter, formula (25) is used wherever the calculation of $f(e)$ is required in theoretical proofs.

Thus, within the ISL theory, two mechanisms for working with anchors are distinguished, solving fundamentally different tasks.

The structural-canonical functional $\varphi$ (Definition 3) answers the question of "*what constitutes an anchor"*: it identifies structurally significant elements by the local topology of the neighborhood, regardless of the comparison context. That is why $\varphi$ is used in the definition of the frequency function (25) and in theoretical proofs: its result depends neither on the order of accumulation nor on the matching algorithm.

The matching operator $\mu$ answers the question of "*which candidate anchor of a new example corresponds to which attractor anchor*": it establishes a correspondence between two specific structures during the accumulation process $\oplus$ (discussed below) and inference. Its result depends on the structures being compared, so it is not used in theoretical proofs, but is necessary for the operational implementation of learning.

3.3. Full Reduction Operator:

$$R = R_s \circ R_p \quad (26).$$

During the reduction process, parametric reduction is performed first, followed by structural reduction.

***Axiom 2.*** *Strict reductivity and minimality.*

$$R(\mathcal{G}) \preceq \mathcal{G} \quad (27).$$

The size of the graph in terms of the number of elements and parameters is strictly non-increasing, and the final attractor is minimal in the sense that it contains no elements with $f(e) < 1$.

**4. Learning Dynamics (accumulative, order-invariant).**

Upon receiving a new positive example ($x'$), is constructed:

$$\mathcal{G}' = D(x') \quad (28).$$

4.1. Anchor matching operator:

$$\mu: E' \supseteq A' \rightarrow E \quad (29),$$

is a partial mapping that correlates the anchors of the new example ($A'$) with existing ones.

The mappings $\mu$ are consistent if, during sequential matching of anchors from different examples, the result does not depend on the matching order, i.e., there are no contradictory identifications of the same anchors.

4.2. Cumulative Union Operation.

Let us have a stored hypergraph $\mathcal{G}_t$ and a new example $\mathcal{G}'$:

$$\mathcal{G}_t = \left(E_t, \mathcal{R}_t^{(s)}, \mathcal{R}_t^{(p)}\right) \quad (30),$$

$$\mathcal{G}' = \left(E', \mathcal{R}'^{(s)}, \mathcal{R}'^{(p)}\right) \quad (31),$$

$$\mu: E' \supseteq A' \rightarrow E_t \quad (32).$$

1. Perform vertex union:

The union $E_{\oplus}$ is defined as follows:

– if $e' \in \mathrm{dom}(\mu)$, then element $e'$ is identified with $\mu(e')$ and is not added as new (i.e., it is replaced by $\mu(e')$ in all edges);

– if $e' \notin \mathrm{dom}(\mu)$, then it is added to $E_t$.

2. Execute edge transfer:

Let us define the mapping:

$$\tilde{\mu}: E' \to E_{\oplus},$$
$$\tilde{\mu}(e') = \begin{cases} \mu(e'), & e' \in \operatorname{dom}(\mu) \\ e', & otherwise \end{cases} \quad (33),$$

then edges are transferred as:

– spatial:

$$\mathcal{R}_{\oplus}^{(s)} = \mathcal{R}_{t}^{(s)} \cup \tilde{\mu}\left(\mathcal{R}'^{(s)}\right) \quad (34),$$

– parametric:

$$\mathcal{R}_{\oplus}^{(p)} = \mathcal{R}_{t}^{(p)} \cup \tilde{\mu}\left(\mathcal{R}'^{(p)}\right) \quad (35).$$

The mapping $\tilde{\mu}$ extends to edges component-wise, i.e., for a hyperedge $r' = \{e'_1, \dots, e'_k\}$ we have:

$$\tilde{\mu}(r') = \{\tilde{\mu}(e'_1), \dots, \tilde{\mu}(e'_k)\} \quad (36).$$

3. Final formula:

$$\mathcal{G}_t \oplus \mathcal{G}' = (E_t \cup_{\mu} E', \mathcal{R}_t^{(s)} \cup \tilde{\mu}(\mathcal{R}'^{(s)}), \mathcal{R}_t^{(p)} \cup \tilde{\mu}(\mathcal{R}'^{(p)})) \quad (37),$$

where:

– $\cup_{\mu}$ denotes union with the identification of elements according to $\mu$;

– $\oplus$ is the cumulative union operation, where identified anchors are glued together, and the remaining hypergraph elements are added.

4.3. Learning Iteration:

$$\mathcal{G}_{t+1} = R(\mathcal{G}_t \oplus \mathcal{G}') \quad (38).$$

Thus, learning is defined as the iteration of an operator on the space of hypergraphs, where the operation $\oplus$ expands the structure, and the operator $R$ performs its reduction.

4.4. Associativity and Commutativity of Learning (up to isomorphism):

$$(\mathcal{G}_1 \oplus \mathcal{G}_2) \oplus \mathcal{G}_3 \sim \mathcal{G}_1 \oplus (\mathcal{G}_2 \oplus \mathcal{G}_3) \quad (39),$$

$$\mathcal{G}_1 \oplus \mathcal{G}_2 \sim \mathcal{G}_2 \oplus \mathcal{G}_1 \quad (40),$$

where $\sim$ is the structural equivalence relation: two hypergraphs are equivalent if their segmented parameters match and their minimal structures are locally isomorphic around common anchors.

Indeed, relations (39) and (40) hold under the consistency of the mappings $\mu$, i.e., in the absence of contradictory identifications of the same anchors, and with a fixed selection rule for $\mu$. The set of anchors $A$ forming the structure $\mathcal{G}_i$ is determined by the frequency function $f(e)$, which is a sum over a fixed set $\mathcal{C}$ (formula 25) and does not depend on the order of summation.

The associativity and commutativity of learning imply that the result of reducing an accumulated finite set of class examples $\mathcal{C}$ does not depend on the order of their sequential presentation, up to isomorphism. Indeed, assuming that each class example $\mathcal{C}$ contains the same hidden hypergraph (up to isomorphism) that manifests during the reduction process, accumulation may alter the trajectory of intermediate states, but not the resulting hidden hypergraph.

Therefore, different presentation orders lead to the same accumulated configuration $\mathcal{G}^*$ of class $\mathcal{C}$ up to isomorphism.

Then we define that, in the general case, two hypergraphs are structurally equivalent:

$$\mathcal{G}_i \sim \mathcal{G}_j \Leftrightarrow \exists \phi : \mathcal{G}_i \to \mathcal{G}_j \quad (41),$$

if there exists a local isomorphism preserving anchors on the subgraphs induced by the anchors.

4.5. Training Exclusively on Positive Examples.

The proposed learning approach provides for the independent training of individual class-detector neurons. In this case, we do not construct hypersurfaces separating classes, as occurs in statistical models; rather, we form an invariant hypergraph structure for all examples of a single class only.

Consequently, training must be carried out exclusively on positive examples of that class.

We hypothesize that the learning process of a class-detector neuron models information processing within the dendritic tree of a single pyramidal neuron. Thus, algebraic reduction of hypergraphs can be interpreted as the structural plasticity of active dendrites, where patterns of synaptic inputs on dendritic branches form complex interrelations of elements that constitute structural invariants [8].

This learning approach can be formulated as the following axiom, which in this formulation has an architectural rather than purely mathematical status.

***Axiom 3.*** *Training of class-detector neurons is performed exclusively on positive examples of the given class.*

Thus, upon receiving input signals, the system does not "compute" an answer, but *evolves over time until it reaches a stable state*. This evolutionary process consists in the system changing its state based on measurements of its own structural components.

The system's dynamics are determined by a function of the set of measured state parameters, which makes it possible to model the internal consistency of the structure without introducing an external optimality criterion.

All subsequent propositions are derived from the axioms defining the properties of the reduction operator and the representation structure.

**5. Principle of Invariant Reduction.**

The system evolves in the direction of reducing structural redundancy while preserving invariant elements until it reaches a minimal fixed structure. This principle does not rely on gradient-based optimization or classical variational principles. Instead, learning is regulated by a monotonic reduction operator in a partially ordered space of structures, converging to minimal invariant fixed points.

Axiom 1 of segmentation defines the discrete structure of the parametric space in which the operator $R_p$ is defined and the directionality of reduction is induced.

1. Let us introduce a segmentation hierarchy of the coordinate system and measurement scales that is uniform across all classes. This is exogenous segmentation, modeling the process of segment self-organization in biological neural structures:

$$\Sigma_i^0 \succ \Sigma_i^1 \succ \cdots \succ \Sigma_i^{L_i} \quad (42),$$

where $\Sigma_i$ are segments of parameter $i$; the larger the index, the more generalized ("coarser") the segments.

2. Let us introduce transition operators between levels:

$$\pi_i^{\ell\to\ell+1}: \Sigma_i^\ell \to \Sigma_i^{\ell+1} \quad (43),$$

where quantization at level $\ell$ is:

$$Q_i^\ell: M_i \to \Sigma_i^\ell, Q^\ell\big(m(\mathcal{G})\big) = \big\{Q_i^\ell\big(m_i(e)\big)\big\} \quad (44).$$

3. Let us define a partial order (poset) on $\mathcal{Z}$ [9]:

$$\mathcal{G}_1 \preceq \mathcal{G}_2 \Leftrightarrow \begin{cases} E_1 \subseteq E_2 \\ \mathcal{R}^{1(s)} \subseteq \mathcal{R}^{2(s)} \ (\textit{or does not introduce any new connectivity}) \\ \Sigma_i^{(\ell_1)} \succeq \Sigma_i^{(\ell_2)} \text{if } \ell_1 \geq \ell_2 \end{cases} \quad (45),$$

where $\preceq$ is a partial order: reflexivity is obvious, and transitivity follows from the transitivity of inclusion and the order relation on the hierarchical structure of segments.

4. Let us define the invariance functional $\mathrm{Inv}(\mathcal{G})$:

$$\mathrm{Inv}(\mathcal{G}) = \{e \in E : f(e) = 1\} \quad (46),$$

which specifies a closure and, in principle, allows for an analogy with the closure operator in Formal Concept Analysis (FCA) [10].

5. Expressing the reduction operators via segments:

Parametric reduction (transition to a coarser level):

$$R_p^{\ell}(\mathcal{G}) = \left(E, \mathcal{R}^{(s)}, Q^{\ell}\big(m(\mathcal{G})\big)\right) \quad (47),$$

with step $\ell \to \ell + 1$ via $\pi^{\ell \to \ell+1}$ (for example, for a coordinate system).

Then, structural reduction can be represented as:

$$R_s(\mathcal{G}) = \left(S(E), closure\big(S(E)\big), \mathcal{R}'^{(p)}\right);\ S(E) = \mathrm{Inv}(\mathcal{G}) \quad (48).$$

Classical neural networks lose the topological structure of an image, whereas ISL models an approach to solving the classical neurobiological Binding Problem through the preservation of anchor connectivities (hyperedges) within the attractor [11].

6. Let us define the discrete invariance gradient, or more precisely, a gradient-like direction, as a geometrically induced segmentation vector that sets the direction for increasing the level of generalization in the segmentation hierarchy $\{\Sigma_i^\ell\}$:

$$\nabla_\Sigma : \mathcal{G} \to R_p^{\ell\to\ell+1}(\mathcal{G}) \quad (49).$$

This gradient has the property (after $R_s$):

$$\mathcal{G} \overset{\nabla_\Sigma}{\to} \mathcal{G}' \preceq \mathcal{G} \quad (50),$$

$$\mathrm{Inv}(\mathcal{G}') \subseteq \mathrm{Inv}(\mathcal{G}) \cup New(\mathcal{G}'), \mathrm{Inv}(\mathcal{G}') \supseteq \mathrm{Inv}(\mathcal{G}) \cap Stable(\mathcal{G}') \quad (51),$$

where:

- $New(\mathcal{G}')$ - elements that first appeared in all observations before time $t + 1$;
- $Stable(\mathcal{G}')$ - elements maintaining frequency $f(e) = 1$ after adding a new example.

The meaning of formula (46) is that invariants can be added or removed (if false), but true invariants are preserved.

Thus, *a non-differential, but order-induced gradient-like dynamics* is created. In general, the dynamics of invariants is not strictly monotonic: elements can both appear and disappear depending on the refinement of the frequency estimate. However, the sequence of invariant sets converges to a stable core of the structure $E^*$, determined by the class.

Then, we can formalize *the principle of invariant reduction with a gradient*:

$$
\begin{array}{|cl|}
\hline
\text{Evolution:} & \mathcal{G}_{t+1} = R(\mathcal{G}_t \oplus \mathcal{G}') \\
\text{Monotony of structure:} & \mathcal{G}_{t+1} \preceq \mathcal{G}_t \oplus \mathcal{G}' \\
\text{Dynamics of invariants:} & \text{Inv}(\mathcal{G}_{t+1}) = \{e: f_{t+1}(e) = 1\} \\
\text{Convergence:} & \exists\, E^*: \text{Inv}(\mathcal{G}_t) \to E^* \\
\text{Gradient:} & R_p\ realizes\ \nabla_\Sigma\ (transition\ \ell \to \ell + 1) \\
\text{Limit:} & \exists\, \mathcal{A} \in \mathcal{Z}: R(\mathcal{A}) = \mathcal{A} \\
\hline
\end{array}
\quad (52).
$$

The features of this principle are:

- the direction is set by the global segmentation lattice $\{\Sigma_i^\ell\}$;
- reduction increases invariance and decreases complexity;
- the fixed point $\mathcal{A}$ is *the structural attractor*.

In this approach, *extremality is induced by the operator and the order rather than by an external optimization functional* (such as the loss function in statistical ANN models):

$$
\mathcal{A} = Fix(R) \Leftrightarrow \begin{cases} \text{Inv}(\mathcal{A}) = E^* \\ R(\mathcal{A}) = \mathcal{A} \\ \forall \mathcal{G}: \text{Inv}(\mathcal{G}) = E^* \Rightarrow \mathcal{A} \preceq \mathcal{G} \end{cases} \quad (53).
$$

The segmentation hierarchy $\Sigma_i$ and the invariance gradient $\nabla_\Sigma$ can be viewed as elements of a mathematical model of "neural syntax" (assembly rules) used by brain neurons to encode information, where the "code" is the hypergraph structure itself [12].

## 6. Structural Attractor and Concept.

The central concept of this theory is the structural attractor - an invariant structure that emerges as a fixed point of the reduction operator and defines the internal model of an object class.

Unlike classical dynamical systems [13, 14], where an attractor is defined as a set of states, in this approach the attractor is a canonical structure - a minimal hypergraph that is stable with respect to the reduction operator.

***Definition 4.*** *Structural attractor of a class.*

Let $\mathcal{C} \subset X$ be a class of objects, $Z$ be the space of hypergraphs, and $R: Z \to Z$ be the reduction operator.

*The structural attractor of class* $\mathcal{C}$ is a hypergraph:

$$\mathcal{A}_{\mathcal{C}} \in Z \quad (54),$$

satisfying the conditions:

1. Invariance:

$$R(\mathcal{A}_{\mathcal{C}}) = \mathcal{A}_{\mathcal{C}} \quad (55),$$

2. Structural minimality:

$$\mathcal{A}_{\mathcal{C}} = \min_{\preceq}\{\mathcal{G} \in Z: \mathrm{Inv}(\mathcal{G}) = E^*\} \quad (56),$$

where:

- $\preceq$ is a partial order on $Z$;
- $E^*$ is the set of true structural invariants of the class:

$$E^* = \bigcap_{x \in \mathcal{C}} H_x \quad (57),$$

where $H_x = D(x)$ are the hypergraphs of class examples, and the intersection of hypergraphs can be viewed as the intersection of vertex sets after alignment.

3. Attraction (convergence of reduction):

For any representation $D(x)$, $x \in \mathcal{C}$, there exists a finite number of steps $T$, such that:

$$R^T\big(D(x)\big) \sim \mathcal{A}_{\mathcal{C}} \qquad (58),$$

where $\sim$ is the structural equivalence relation.

Due to the fixed-point property of $\mathcal{A}_{\mathcal{C}}$, this property entails the stabilization of iterations.

This property means that any structure of the class transitions into the structural attractor in a finite number of reduction steps, and the reduction operator acts as an *attraction* to the $\mathcal{A}_{\mathcal{C}}$.

The structural attractor $\mathcal{A}_{\mathcal{C}}$ is:

- a fixed point of the reduction operator;
- a minimal carrier of the invariant structure;
- the result of the convergence of the learning process;
- a canonical form for representing objects of the class.

Thus, learning in this model represents the process of reducing various object representations to a single structural normal form.

***Definition 5.*** *Concept.*

A *Concept* is a structural attractor of a perceived holistic image that serves as its internal model, meaning the Concept is an endogenous ontological object for the given system:

$$\text{Concept } \mathcal{C} \equiv \mathcal{A}_{\mathcal{C}} \qquad (59).$$

The formation of structural attractors makes it possible to isolate such attractor systems/models into a distinct class of *dynamic self-organizing systems equipped with* *memory of structural attractors*. Biological neurons and neural structures serve as one example of such systems.

The proposed model admits a neurobiological interpretation, wherein:

- the structural attractor corresponds to a stable activation pattern;
- anchor elements are realized through stable synaptic configurations;
- reduction reflects the selection of stable structures within dendritic trees.

This interpretation is of a model-based nature and is not a direct assertion regarding biological implementation.

Self-organization in this context refers not to a physical process, but to a modeled one that entails specific modeling constraints - in particular, the exogenous definition of primitive detectors, coordinate systems, and scales. Simultaneously, the reduction operators model the energy dynamics of natural self-organization.

Thus, a structural attractor is a minimal invariant structure that acts as a fixed point of the reduction operator, defining the concept as the internal model of a class.

## 7. Convergence Theorem for the Reduction Operator.

All theoretical proofs are carried out under the condition that each example hypergraph from the training set contains the complete hidden attractor of the class.

***Theorem 1.*** *Convergence of the reduction operator.*

Let $Z$ be the space of hypergraphs equipped with a partial order $\preceq$, and $R: Z \to Z$ be the structural reduction operator.

Let us introduce the complexity functional:

$$C(\mathcal{G}) = |E(\mathcal{G})| + \lambda |\mathcal{R}_{\mathcal{G}}| \quad (60),$$

where:

- $C(\mathcal{G})$ is a measure of complexity (or "size") of the structure that accounts for both structural and parametric complexity;

- $E(\mathcal{G})$ is the set of hypergraph vertices: $E(\mathcal{G}) \in \mathbb{N}$,
- $\mathcal{R}_{\mathcal{G}}$ is the set of its edges (hyperedges): $\mathcal{R}_{\mathcal{G}} \in \mathbb{N}$,
- $\lambda > 0$ is a fixed (empirical) coefficient accounting for the different contributions of vertices and edges to the structure's complexity.

*Remarks:*

1. The complexity functional $C$ can be interpreted as a d*iscrete Lyapunov function* for the reduction dynamics $\mathcal{G}_{t+1} = R(\mathcal{G}_t)$: it is non-increasing along trajectories and strictly decreasing outside of fixed points. Consequently, the reduction process can be viewed from a physical standpoint as dissipative, and the structural attractor corresponds to the minimal (in terms of $C$) fixed structure.

2. The complexity functional $C$ pertains *not to the data accumulation process*, but to the reduction of *a fixed accumulated structure*.

3. From a neurobiological perspective, the complexity functional $C$ is a simplified mathematical abstraction designed to measure the structural and parametric redundancy of a hypergraph relative to its corresponding attractor. The values of this functional decrease during the reduction process and can be used to make decisions regarding class separation during WTA (Winner-Take-All) competition.

Based on Axioms 1 and 2, the reduction operator possesses the following properties:

1. Non-expansion of the structure: $R(\mathcal{G}) \preceq \mathcal{G}, \forall \mathcal{G} \in Z$;
2. Strict decrease in complexity outside of fixed points:

$$R(\mathcal{G}) \neq \mathcal{G} \Rightarrow C\big(R(\mathcal{G})\big) < C(\mathcal{G}) \tag{61}$$

By the definition of the reduction operator, non-invariant elements $E(\mathcal{G})$ and/or $\mathcal{R}_{\mathcal{G}}$ are removed, simplifying the structure, which strictly decreases the complexity $C$.

3. Discreteness of complexity.

*Definition of the Theorem 1.*

Since the number of elements is finite and complexity values are discrete, for any initial hypergraph $\mathcal{G}_0 \in Z$, the sequence of reductions $\mathcal{G}_{t+1} = R(\mathcal{G}_t)$ stabilizes in a finite number of steps:

$$\exists T : \mathcal{G}_T = \mathcal{G}_{T+1} \quad (62),$$

and the resulting structure $\mathcal{G}_T$ is a fixed point $R(\mathcal{G}_T) = \mathcal{G}_T$.

*Proof:*

***Step 1.*** Decrease of the complexity functional.

If $\mathcal{G}_t$ is not a fixed point, then by condition (61):

$$C(\mathcal{G}_{t+1}) = C\big(R(\mathcal{G}_t)\big) < C(\mathcal{G}_t) \quad (63).$$

It is assumed that the reduction operator $R$ induces a strict transition in the partial order $\preceq$, i.e., $R(\mathcal{G}) \preceq \mathcal{G}$ and $R(\mathcal{G}) \neq \mathcal{G} \Rightarrow R(\mathcal{G}) \prec \mathcal{G}$.

Indeed, if at least one of the inclusions in equation (45) is strict, then the complexity functional (60) is strictly compatible with the order $\mathcal{G}_1 \prec \mathcal{G}_2 \Longrightarrow C(\mathcal{G}_1) < C(\mathcal{G}_2)$.

***Step 2.*** Finiteness of the decreasing sequence.

Since $C(\mathcal{G}_t)$ is bounded from below by zero, and the sequence $\{C(\mathcal{G}_t)\}$ of complexities is strictly decreasing, it cannot be infinite.

**Step 3.** Stabilization.

Consequently, there exists a finite moment $T$ starting from which the decrease stops, i.e., $C(\mathcal{G}_T) = C(\mathcal{G}_{T+1})$.

From condition (61), it follows that this is possible only if:

$$R(\mathcal{G}_T) = \mathcal{G}_T \quad (64).$$

Thus, the sequence stabilizes in a finite number of steps, and its limit is a fixed point of the reduction operator. ■

*Remark.*

An estimate for the number of convergence steps can be obtained via the initial complexity:

$$T \leq C(\mathcal{G}_0) \quad (65).$$

Then, if there exists a *unique* structural attractor $\mathcal{A}_{\mathcal{C}} = Fix(R)$ induced by class $\mathcal{C}$, then for any $x \in \mathcal{C}$:

$$R^t\big(D(x)\big) \to \mathcal{A}_{\mathcal{C}}\ (with\ an\ accuracy\ of\ \sim) \quad (66),$$

that is:

$$\exists T : R^T\big(D(x)\big) \sim \mathcal{A}_{\mathcal{C}} \quad (67).$$

Consequently, we need to prove the uniqueness of the class's structural attractor.

## 8. Theorem on the Uniqueness of the Class Structural Attractor.

***Theorem 2.*** *Uniqueness of the class structural attractor.*

Let the following conditions be satisfied for class $\mathcal{C}$:

1. The accumulated structural configuration of the class is finite;
2. The invariance criterion is fixed;
3. The set of anchors is determined by this criterion;
4. Local structural reduction removes non-anchor elements;

5. Closure is a fixed deterministic operator for restoring connectivity on a given base spatial graph;

6. The base spatial structure $\mathcal{R}^{(p)}$ is fixed.

Then the full structural reduction operator $R_s = G \circ L$ has a unique fixed point $\mathcal{G}_{\mathcal{C}}^*$ reached from a given accumulated configuration of the class.

In other words:

$$R_s(\mathcal{G}_{\mathcal{C}}^*) = \mathcal{G}_{\mathcal{C}}^* \qquad (68).$$

*Proof:*

***Step 1.*** By Lemma 1, local structural reduction achieves the stabilized set of anchors $A$ in a finite number of steps. The uniqueness of the anchor set is determined by the accumulated structure of the class.

After removing non-anchor elements, Closure is applied:

$$\mathcal{G}_{\mathcal{C}}^* = G\left(A, \mathcal{R}^{(p)}\right) \qquad (69).$$

For fixed $A$ and $\mathcal{R}^{(p)}$, the result of Closure is unique (due to the determinism property of the Closure operator). Consequently, there exists a unique structure $\mathcal{G}_{\mathcal{C}}^*$ that emerges after reduction stabilization and subsequent spatial closure.

***Step 2.*** In accordance with the idempotency property of the Closure operator, its repeated application does not alter the resulting structure. Since all remaining elements are anchors after stabilization, repeated local reduction also removes no elements. Consequently:

$$R_s(\mathcal{G}_{\mathcal{C}}^*) = (G \circ L)(\mathcal{G}_{\mathcal{C}}^*) = \mathcal{G}_{\mathcal{C}}^* \qquad (70).$$

Thus, $\mathcal{G}_{\mathcal{C}}^*$ is a fixed point of the full structural reduction operator.

***Step 3.*** Suppose now that there exist two fixed points $\mathcal{G}_1^* \neq \mathcal{G}_2^*$. Since both are the result of stabilizing the same accumulated class configuration under the same invariance criterion, by the uniqueness property of the anchor set, both must have the same set of anchors: $A_1 = A_2 = A$.

Since the base spatial structure $\mathcal{R}^{(p)}$ is also fixed, in accordance with the determinism property of the Closure operator:

$$\mathcal{G}_1^* = \mathrm{Closure}(A) = \mathcal{G}_2^* \quad (71),$$

i.e., $\mathcal{G}_1^* = \mathcal{G}_2^*$, which contradicts the assumption. Consequently, the fixed point is unique:

$$\exists!, \mathcal{G}_{\mathcal{C}}^*: R_s(\mathcal{G}_{\mathcal{C}}^*) = \mathcal{G}_{\mathcal{C}}^* \quad (72).$$

■

***Corollary 1.*** *Conditional uniqueness of the class structural attractor.*

Theorem 2 does not assert the global uniqueness of the structural attractor across the entire space of possible structures.

It asserts intra-class uniqueness under fixed conditions for the formation of a given class: $\mathcal{C}$, $A$, invariance criteria, and $\mathcal{R}^{(p)}$. Therefore, for two different classes $\mathcal{C}_1$ and $\mathcal{C}_2$, it is not strictly required that $\mathcal{G}_{\mathcal{C}_1}^* \neq \mathcal{G}_{\mathcal{C}_2}^*$.

In particular,

$$\mathcal{C}_1 \neq \mathcal{C}_2 \nRightarrow \mathcal{G}_{\mathcal{C}_1}^* \neq \mathcal{G}_{\mathcal{C}_2}^* \quad (73),$$

since the same structural configuration can acquire different meanings in different contexts.

Consequently, an essential property of a biological system is not the global uniqueness of attractors, but *the distinguishability of structural attractors within the contextual space available to the system*.

## 9. Parametric Spaces of Hypergraphs and Their Separability.

For the practical implementation of the separability of structural attractors, let us distinguish between two parametric spaces of a hypergraph:

- *Relative (structure-centric) parametric space:* A space in which the parameters of each structural element are considered locally - the coordinate system is placed independently at each structural point under consideration. It is in this space that the local reduction $R_p$ operates and anchors $A$ are formed.
- *Absolute parametric space:* A space in which the parameters of the entire structure are considered relative to a single fixed coordinate system whose origin is fixed relative to the whole structure (e.g., placed at the center of mass of the foreground). Characteristics of the structure as a whole are determined in this space.

***Definition 6.*** *A global parametric invariant of class $\mathcal{C}$ is a functional:*

$$\Phi: Z \to \Gamma = \Gamma_1 \times \Gamma_2 \times \cdots \times \Gamma_k, \quad (74),$$

where each $\Gamma_i$ is a finite set of allowed values for the $i$-th global parametric invariant, satisfying the following conditions:

1. *Constancy on the class:*

$$\forall x \in \mathcal{C}: \Phi(H_x) = \Phi_{\mathcal{C}} \in \Gamma \quad (75),$$

2. *Determinism:* $\Phi$ is computed unambiguously from the topological structure of the hypergraph and does not depend on the order of example accumulation.

*Note:* The determinism condition is satisfied trivially, as $\Phi$ is computed from the structure of a single hypergraph $H_x$, which is independent of the accumulation order of other examples. Constancy $\Phi_C$ on the class (Condition 1) is a property of class $C$ rather than a result of the accumulation procedure.

*3. Resistance to reduction:*

$$\Phi(R(\Gamma)) = \Phi(\Gamma) \quad (76),$$

that is, the reduction operator preserves global parametric invariants.

Examples of global parametric invariants for skeletonized MNIST contours include:

- Closedness or disconnectedness of the entire structure or individual substructures; monotonicity or non-monotonicity of parameter changes in structural elements during sequential contour traversal (e.g., monotonic change of local segment orientation defines the structure's convexity; direction of parameter changes for segments or angles), etc.;
- Contour traversal direction (clockwise or counterclockwise);
- Stable positions of individual structural elements or substructures within the absolute coordinate system.

It is precisely the global parametric invariants that can *separate* classes possessing identical local anchors and spatial interconnections. For example, the digits "6" and "9" have identical local anchors and connection topologies, but differ in their global parametric invariants: the winding directions of the contour relative to a single endpoint and the position (orientation) of this endpoint in the absolute coordinate system.

Note that global parametric invariants $\Phi_{\mathcal{C}}$ in this model also have an exogenous nature and can be interpreted as *global parametric anchors* that preserve the spatial topology of the reduced hypergraph analogously to the original one.

Then, the combination of local anchors $A$ and global parametric invariants $\Phi_{\mathcal{C}}$ forms the *full invariant signature of the class:*

$$\Psi_{\mathcal{C}} = (A_{\mathcal{C}}, \Phi_{\mathcal{C}}) \quad (77).$$

We can conclude that for any two classes $\mathcal{C}_i \neq \mathcal{C}_j$, these classes must be distinguishable by their full invariant signature:

$$\Psi_{\mathcal{C}_i} \neq \Psi_{\mathcal{C}_j} \quad (78).$$

It is the full invariant signatures of classes that serve as the basis for the separability (*distinguishability*) of their structural attractors. However, as experimental studies presented below have shown, a key challenge lies in constructing this signature - specifically, its implementation in the form of an edit distance on hypergraphs.

**10. Attractor as a "Fuzzy" Prototype of a Class.**

The attractor $\mathcal{A}_{\mathcal{C}}$ = ($E^*$, $\sigma^*$, $\Sigma^*$) is unique as a structural object, but the parametric fiber bundle $\Sigma^*$ contains intervals and qualitative categories rather than point values. This means that the attractor specifies not a single concrete structure, but a class of equivalent realizations:

$$[\mathcal{A}_{\mathcal{C}}] = \{\mathcal{G} \in Z \mid E(\mathcal{G}) = E^*, \sigma(\mathcal{G}) = \sigma^*, \forall e \in E^*: Q(m(e)) \in \Sigma_e^*\} \quad (79).$$

Each element in $[\mathcal{A}_\mathcal{C}]$ is a valid structurally identical, but parametrically distinct, instantiation of the attractor. Upon visualization, we sample one element from this class each time and obtain different structures belonging to the same class.

***Definition 7.*** *Class of attractor realizations.*

*The class of realizations* of the structural attractor $\mathcal{A}_\mathcal{C}$ is called the set defined above, where each $\mathcal{G}$ is structurally equivalent to $\mathcal{A}_\mathcal{C}$.

These considerations can be formalized as a corollary.

***Corollary 2.*** The attractor $\mathcal{A}_\mathcal{C}$ is unique as a structural object, but generates a continuum of parametric realizations $[\mathcal{A}_\mathcal{C}]$. At the same time:

1. Any class example is a realization of the attractor: $\forall x \in \mathcal{C}: D(x) \in [\mathcal{A}_\mathcal{C}]$;
2. The width of the parametric intervals $\Sigma^*$ reflects intra-class variability;
3. Augmentation expands $[\mathcal{A}_\mathcal{C}]$ without changing $\mathcal{A}_\mathcal{C}$ as a structural object.

This corollary admits a neurobiological interpretation:

1. Larkum (2022) [8] and Branco & Häusser (2010) [15] show that dendritic branches function as local pattern comparators. Crucially, they respond not to the exact reproduction of a pattern, but to falling within the zone of permissible values - that is, specifically to belonging to $[\mathcal{A}_\mathcal{C}]$, rather than an "exact" match with $\mathcal{A}_\mathcal{C}$. The parametric intervals $\Sigma^*$ in this model can be viewed as a formal analogue of this zone of permissible synaptic inputs.

2. There is a neurobiological hypothesis, tracing back to Hebb's work and developed in the concept of "sparse distributed representation" (Kanerva (1988) [16]), that the brain stores not exact copies of examples, but distributed representations from which an approximate realization is reconstructed upon retrieval. This explains a well-known psychological fact: memories are not exact copies; they are reconstructed anew each time and may vary slightly. In this model, this is formalized by storing $\mathcal{A}_\mathcal{C}$ while reproducing an element from $[\mathcal{A}_\mathcal{C}]$.

3. This also echoes the concept of prototype representation in cognitive psychology by Rosch (1975) [17]: people classify objects not by an exact match with a template, but by proximity to a prototype - an abstract generalized representation of a category that does not coincide with any single specific example. The attractor $\mathcal{A}_{\mathcal{C}}$ can be viewed as a formal prototype in this sense: it is the structural invariant of the class, whereas the class of realizations $[\mathcal{A}_{\mathcal{C}}]$ forms the region of permissible proximity around it.

In classical machine learning models, class variability is a function of an external evaluation: it is defined via a loss function and the statistical distribution of the training set. The model represents variability only to the extent that an external criterion penalizes its omission.

In the proposed model, variability arises endogenously. The parametric intervals $\Sigma^*$ are formed directly in the process of cumulative reduction: the width of the interval for each parameter of each anchor $e \in E^*$ is an exact reflection of the observed scatter of this parameter across all class examples. No external criterion is required for this.

Thus, the attractor $\mathcal{A}_{\mathcal{C}}$ encodes two objects simultaneously: the structural invariant of the class (topology $E^*, \sigma^*$) and the endogenously formed measure of intra-class variability ($\Sigma^*$). This fundamentally distinguishes ISL from metric learning methods, where the metric is defined or learned exogenously.

Taking into account model constraints (exogeneity of primitive detectors, coordinate systems, and scales), this statement holds true for the parametric layer of the model. Fully endogenous variability, including the formation of coordinate systems and primitives, remains an area for future research.

**11. Few-Shot Learning.**

Let us consider a subset $\mathcal{S} \subset \mathcal{C}$. We are interested in the conditions under which training on $\mathcal{S}$ leads to the same attractor as training on the entire class $\mathcal{C}$.

*Coverage Condition.*

Let:

$$\mathcal{A}_{\mathcal{C}} \subseteq \bigcup_{x \in \mathcal{S}} H_x \qquad (80),$$

i.e., all elements of the attractor are present in at least one example from $\mathcal{S}$.

***Definition 8.*** *Sufficient covering set.*

A subset $\mathcal{S} \subset \mathcal{C}$ is called *a sufficient covering set* if the following conditions are met:

1. *Coverage* (80);
2. *Anchor connectivity*: the anchor overlap hypergraph is connected, meaning that the structures of all examples from $\mathcal{S}$ can be aligned via common anchors.

By *anchor connectivity*, we mean the set of structural relations between anchors that are induced by the initial spatial structure of the accumulated configuration and are preserved when non-anchor elements are removed.

***Corollary 3.*** If $\mathcal{S}$ is a sufficient covering set, then:

$$\lim_{t \to |\mathcal{S}|} \mathcal{G}_t(\mathcal{C}) \preceq \mathcal{A}_{\mathcal{C}} \qquad (81),$$

and all attractor elements are preserved during reduction.

*Proof:*

**Step 1.** *Representing the attractor via a subset.*

Here we return once again to matching: $\mu$ is used to establish correspondence between structures rather than to define anchors.

Under the consistency of mappings $\mu_x$:

$$\mathcal{A}_{\mathcal{C}} = \bigcap_{x \in \mathcal{S}} \mu_x\left(\mathrm{Inv}(H_x)\right) \qquad (82),$$

if $\mathcal{S}$ contains all invariant elements.

***Step 2.*** *Defining the nesting property.*

Let:

$$\mathcal{A}_{\mathcal{C}} = \left(E^*, \mathcal{R}^{(s)*}, \mathcal{R}^{(p)*}\right) \quad (83).$$

Then for any $x \in \mathcal{C}$, there exists an embedding mapping:

$$\mu_x : E^* \to E_x \quad (84),$$

such that:

- $\mu_x(E^*) \subseteq E_x$;
- structural relations are preserved;

i.e., the attractor is nested within each example.

Thus, the attractor can be determined not by all examples of the class, but merely by a small subset of its carriers (the entire set of class examples may be redundant). This corresponds to human perceptual psychology and paves the way for few-shot learning.

*Therefore, the attractor is completely determined by the set of its anchors and their connections, while the learning process becomes the reconstruction of its hidden structure from partial observation.* ■

*Note:* A sufficient covering set ensures the presence of all structural elements of the attractor in the sample (reconstructibility of the attractor structure), but does not guarantee its uniqueness.

***Definition 9.*** *Identifying set.*

A subset $\mathcal{S} \subset \mathcal{C}$ is called an *identifying set* if the following conditions are met:

1. *Element coverage*:

$$\forall e \in E^* \ \exists x \in \mathcal{S}: \mu_x(e) \in H_x \quad (85).$$

2. *Connection coverage*:

$$\forall r \in \mathcal{R}^* \ \exists x \in \mathcal{S}: \mu_x(r) \subseteq H_x \quad (86).$$

3. *Anchor connectivity* (Definition 8).

Thus, $\mathcal{S}$ is *the minimal coverage of the attractor* sufficient to fully reconstruct the attractor as a structure, meaning this set induces a unique attractor if:

$$\nexists \mathcal{S}' \subset \mathcal{S}: \mathcal{S}' \text{ identifies } \mathcal{A}_{\mathcal{C}} \quad (87).$$

Unlike a sufficient set, an identifying set is defined as a subset *that uniquely reconstructs the structural attractor* - i.e., it induces a unique fixed point of the reduction operator.

Any identifying set is a sufficient covering set, but the converse is generally false.

Therefore, if $\mathcal{S}$ is a minimal covering set, the proof of Theorem 2 is applicable to $\mathcal{S}$.

***Corollary 4.*** Attractor reconstruction.

If:

- the attractor is nested in each example;
- there exists an identifying set $\mathcal{S}$,

then $\mathcal{A}_{\mathcal{C}} = \mathcal{A}_{\mathcal{S}}$ and $\lim \mathcal{G}_t(\mathcal{S}) = \mathcal{A}_{\mathcal{C}}$.

Consequently, what matters is not the cardinality (size) of $|\mathcal{S}|$, but its covering capacity. That is, 2-3 examples might be sufficient, while 100 examples might be insufficient; sufficiency is determined not by the size of the set, but by the fulfillment of the coverage and connectivity conditions.

If the number of anchors is finite and connections are finite, a finite minimal set of examples exists for reconstructing the class attractor.

Thus, we state that the attractor $\mathcal{A}_C$ is *nested within each class example*, and a finite subset $\mathcal{S}$ exists that covers all anchors and connections of the attractor. Consequently, the attractor is fully reconstructed from $\mathcal{S}$.

In this case, data augmentation plays a role fundamentally different from that in the classical approach.

## 12. The Role of Augmentation in Learning.

Augmentation is viewed as *an operator for exploring the parametric space of a structure.*

Let the initial example be given by the hypergraph:

$$H_x = \left(E_x, \mathcal{R}_x^{(s)}, m_x\right) \quad (88).$$

Augmentation generates a family:

$$A(x) = \left\{H_x^{(1)}, \dots, H_x^{(k)}\right\} \quad (89),$$

where:

- structures $H_x^{(j)}$ are equivalent (up to matching);
- parameters $m_x^{(j)}$ vary.

*Parametric Coverage.*

Augmentation is used to construct parameter segmentation:

$$Q_i: M_i \to \Sigma_i \quad (90),$$

based on the set of observed values:

$$\bigcup_{H \in A(x)} \{ m_i(e) : e \in E_x\} \quad (91).$$

It is assumed that parametric variations are independent of the structure - that is, variations obtained from a single example are representative of the entire class in terms of parameters.

*Decomposition of Learning.*

Learning breaks down into two levels:

1. *Structural level.*

Based on the subset $\mathcal{S} \subset \mathcal{C}$, the following is reconstructed:

$$\mathcal{A}_{\mathcal{C}}^{0} = \left(E^{*}, \mathcal{R}^{(s)*}\right) \quad (92),$$

- the structural skeleton (anchors and their connections).

2. *Parametric level.*

Augmentation forms the parametric fiber bundle:

$$\Sigma^{*} = \{\Sigma_e : e \in E^{*}\} \quad (93).$$

The full attractor:

$$\mathcal{A}_{\mathcal{C}} = \left(\mathcal{A}_{\mathcal{C}}^{0}, \Sigma^{*}\right) \quad (94),$$

or

$$\mathcal{A}_{\mathcal{C}} = \left(E^{*}, \mathcal{R}^{(s)*}, \Sigma^{*}\right) \quad (95).$$

Then, if:

- $\mathcal{S}$ covers the structure of the attractor,
- augmentation covers the parametric variations,

then under the consistency of mappings $\mu$:

$$\mathcal{A}_{\mathcal{C}} = \mathcal{A}_{\mathcal{S}} \quad (96).$$

Unlike classical models, where augmentation increases the sample size and reduces overfitting, in this approach augmentation *builds the parametric structure (metric) via segmentation*.

Thus, structural examples determine the topology, while augmentation determines the geometry (metric).

Then, the attractor is reconstructed as a combination of:

- structural coverage (a small $\mathcal{S}$);
- parametric coverage (augmentation).

Augmentation acts as a *metrization operator*.

This allows us to consider the induced *metric on the attractor*:

Let the attractor be given as (95):

$$\mathcal{A}_{\mathcal{C}} = \left(E^*, \mathcal{R}^{(s)*}, \Sigma^*\right),$$

where $\Sigma^* = \{\Sigma_e : e \in E^*\}$ is the parametric fiber bundle.

For each $e_i \in E^*$ in different examples $x$ and $y$, let us define a local metric:

$$d_e(x, y) = d_{\Sigma_e}\left(Q_e\left(m_x(e)\right), Q_e\left(m_y(e)\right)\right) \quad (97).$$

The global metric on the attractor in the simplest case can be defined as:

$$d(x,y) = \frac{1}{|E^*|} \sum_{e \in E^*} d_e(x,y) \qquad (98).$$

Here, an equal contribution of all attractor elements is assumed, since after reduction every element is a structural invariant of the class. However, as experimental studies presented in Section III have shown, such a "straightforward" approach to defining the global metric has a number of critical drawbacks.

*Properties of the Metric:*

1. *Induced by segmentation*: The metric is derived directly from $d \equiv d(\Sigma^*)$;
2. *Invariance*: If $Q_e(m_x(e)) = Q_e(m_y(e))$, then the contribution of the element is zero;
3. *Consistency*: The metric is defined solely on anchors and is consistent through the mapping $\mu$.

Thus, the metric is not specified externally (exogenously), but arises endogenously as a result of learning: *structure* → *segmentation* → *metric*.

This is the opposite of existing paradigms such as metric learning, contrastive learning, and kernel methods.

However, the model has a limitation: the parametric space - which forms the foundation of the metric - is not formed endogenously (as it is in biological systems based on sensory inputs and their projections into neural structures), but is instead specified exogenously. Consequently, we cannot claim a fully endogenous nature of the metric within this proposed model.

**13. Attention Mechanism and Anchors.**

One of the most important problems in implementing the learning process is the hypergraph matching problem, which in its general formulation is NP-hard. We relax the difficulty of solving this problem by using an *attention operator* over structural anchor points.

***Definition 10.*** *Attention operator.*

*The attention operator* (focus of attention) $\mathcal{F}$ is a local element selector:

$$\mathcal{F}(\mathcal{G}) \subseteq E \quad (99),$$

which isolates a subset of elements used to construct matching based on their local informativeness (for example, image contour "capture" points for MNIST).

For example:

$$\mathcal{F}(\mathcal{G}) = \{\, e \in E \mid I(e) \geq \tau_F \,\} \quad (100),$$

where:

- $I(e)$ is the local informativeness measure,
- $\tau_{\mathcal{F}}$ is the threshold.

The attention operator selects only elements with extreme parameter values for a given structure or rarity/contrast (points with maximum informativeness in the Shannon sense).

Instead of searching for global graph isomorphism (Graph Edit Distance, GED) [18], the attention operator restricts matching to a subset $\mathcal{F}(\mathcal{G})$, which translates the problem from global combinatorial optimization into local alignment. That is, even if we do not entirely avoid solving the NP-hard problem of computing exact graph edit distances [19], we significantly weaken its difficulty, which is confirmed by experimental results (Section III).

Thus, matching is defined only on $\mathcal{F}(\mathcal{G})$:

$$\mu : \mathcal{F}(\mathcal{G}') \to \mathcal{F}(\mathcal{G}) \quad (101).$$

The attention operator $\mathcal{F}(\mathcal{G})$ and the structural-canonical functional $\varphi$ (24) are related by:

$$\mathcal{F}(\mathcal{G}, \text{context}) \subseteq \varphi(\mathcal{G}) \qquad (102),$$

that is, $\mathcal{F}(\mathcal{G})$ is a restriction of $\varphi$ (24): the set of elements isolated by the attention operator in a specific context is always a subset of structurally significant points determined canonically by the functional $\varphi$.

At the same time, relation (102) reflects two interrelated processes. First, $\varphi$ specifies a canonical set of points that involuntarily capture the focus of attention regardless of context, which corresponds to the experimental results of Yarbus (1967) [20], who showed that saccades (rapid movements of visual attention focus) are involuntarily attracted to structurally significant points of an image. Second, $\mathcal{F}$ narrows this set based on the current task context.

The functional $\varphi$ is used in the definition of the frequency function (25) and proofs. The operator $\mathcal{F}$ is used in the matching procedure during training and inference. The canonical definition of the operator $\mathcal{F}$ is an open problem of this theory and represents a direction for future research.

We assume that the attention operator can model the process of active perception in the brain [20–22]. In the context of ISL, this can justify the choice of "anchor points" in space as the purposeful work of the attention operator $\mathcal{F}$, minimizing the redundancy of parallel perception.

**14. Neuron Map.**

Class $\mathcal{C}$ is represented as a self-organizing map of attractors:

$$\mathcal{M}_{\mathcal{C}} = \{\mathcal{A}_1, \dots, \mathcal{A}_k\} \quad (103),$$

- a finite set of structural attractors (detector-neurons) corresponding to subclasses or stable variations within the class.

Each attractor $\mathcal{A}_i \in \mathcal{M}_{\mathcal{C}}$ satisfies the condition:

$$\mathcal{A}_{\mathcal{C}} \preceq \mathcal{A}_i \quad (104),$$

that is, the attractor of class $\mathcal{C}$ is a minimal invariant substructure embedded within each subclass attractor or individual example attractor.

The distance between different attractors is bounded from below:

$$d(\mathcal{A}_i, \mathcal{A}_j) > \delta, i \neq j \quad (105),$$

which implies their structural distinguishability.

Consequently, the class neuron map has a strict hierarchy: the central neuron is the class detector neuron - the class concept - and the distance $d$ defines the edit distance of the central neuron (its hypergraph) from the hypergraphs (attractors) of subclass detector-neurons or individual memorized example hypergraphs. The formation of subclass attractors or individual example hypergraphs occurs under special conditions for determining class example novelty, the formulation of which is beyond the scope of this work [1].

***Theorem 3.*** *Conditions for the self-organization of a class attractor map.*

Let:

1. A reduction operator $R$ be given, satisfying the considered properties (monotonicity, invariant selection, connectivity closure);

2. Learning be defined by dynamics (38): $\mathcal{G}_{t+1} = R\left(\mathcal{G}_t \oplus H_{x_{\pi(t)}}\right)$; for each class $\mathcal{C}$, a structural attractor exists (Theorem 2): $\mathcal{A}_{\mathcal{C}} = \{ e \mid f(e) = 1 \}$;
3. A distance $d$ on the space of structures be defined as a pseudometric up to structural equivalence;
4. A distinguishability threshold $\delta > 0$ be fixed.

Then, during the learning process, *a finite map of attractors* $\mathcal{M}_{\mathcal{C}} = \{\mathcal{A}_1, \dots, \mathcal{A}_k\}$ *is endogenously formed* such that the following hold:

1. Structural nesting:

$$\mathcal{A}_{\mathcal{C}} \subseteq \mathcal{A}_i, \forall i \tag{106}$$

2. Separability via formula (105): $d(\mathcal{A}_i, \mathcal{A}_j) > \delta, i \neq j$;
3. Attraction: for any example $x \in \mathcal{C}$:

$$\lim_{t \to T} R^t(H_x) \sim \mathcal{A}_i \text{ for some } \mathcal{A}_i \in \mathcal{M}_{\mathcal{C}} \tag{107}$$

4. Self-organization: the set $\mathcal{M}_{\mathcal{C}}$ is not predetermined, but emerges as a set of stable fixed points of the learning dynamics;
5. Finiteness:

$$\mid \mathcal{M}_{\mathcal{C}} \mid < \infty \tag{108}$$

*Proof:*

***Step 1.*** *Convergence to fixed points.*

By Theorem 1: $\exists T: \mathcal{G}_T = R(\mathcal{G}_T)$, meaning any learning trajectory stabilizes at a fixed point of the operator *R*.

Consequently, all limiting structures have the form (53, 67): $\mathcal{A}_i = \mathrm{Fix}(R)$.

***Step 2.*** *Nature of the fixed points.*

By definition (15): $\mathcal{A} = \{ e \mid f(e) = 1 \}$. This implies:

- each fixed point is determined by *the invariants of a certain subset of examples*,
- different subsets can induce different attractors.

***Step 3.*** *Mechanism for the emergence of new attractors.*

Consider the current attractor $\mathcal{A}_i$ and a new example $H_x$.

Two cases are possible:

1. Consistent case:

$$d(H_x, \mathcal{A}_i) \leq \delta \quad (109).$$

Then after accumulation and reduction:

$$R(\mathcal{A}_i \oplus H_x) \sim \mathcal{A}_i \quad (110),$$

and the structure remains unchanged;

2. Inconsistent case:

$$d(H_x, \mathcal{A}_i) > \delta \quad (111).$$

Then:

- some elements do not match $\mathcal{A}_i$,
- reduction cannot lead to $\mathcal{A}_i$,

and *a new fixed point arises*:

$$R^t(H_x) \to \mathcal{A}_j, \qquad \mathcal{A}_j \nsim \mathcal{A}_i \quad (112).$$

***Step 4.*** *Endogenous map formation.*

Repeating the process for a sequence of examples, we obtain a set of fixed points (103): $\mathcal{M}_{\mathcal{C}} = \{\mathcal{A}_1, \dots, \mathcal{A}_k\}$, where:

- each $\mathcal{A}_i$ arises as the limit of some trajectory,
- no attractor is specified externally.

Thus, the maps self-organize: the structure $\mathcal{M}_{\mathcal{C}}$ is induced by dynamics rather than being set a priori.

***Step 5.*** *Structural nesting.*

By Theorem 2: $\mathcal{A}_{\mathcal{C}} = \{\, e \mid f(e) = 1 \text{ all over } \mathcal{C} \,\}$. Consequently, for any subset: $\mathcal{A}_{\mathcal{C}} \subseteq \mathcal{A}_i$, i.e. the attractor of a class is a common substructure of all attractors of the map.

***Step 6.*** *Separability.*

If distinguishability above the threshold $\delta > 0$ exists between attractors, then $d(\mathcal{A}_i, \mathcal{A}_j) > \delta$ (105), which guarantees that:

- attractors correspond to different stable configurations,
- they do not merge during reduction.

***Step 7.*** *Finiteness.*

The space of structures is finite due to boundedness: the number of elements and parameter segmentations.

Consequently, the number of fixed points is finite (103): $\mid \mathcal{M}_{\mathcal{C}} \mid < \infty$ .

The attractor map emerges as a set of stable fixed points of the dynamics, and its structure is determined by the reduction operator $R$, the accumulation mechanism $\oplus$, and the distinguishability metric $d$, rather than being specified externally:

$$\mathcal{M}_{\mathcal{C}} = \mathrm{Fix}(R), \text{ induced by learning dynamics} \qquad (113).$$

■

Thus, the map is the result of a dynamic partitioning of the example space into regions of attraction.

Naturally, the question arises as to who determines the distinguishability threshold $\delta$ and initiates the formation of a new attractor on the class map.

Theorem 3 establishes the structural condition for creating a new attractor, but the mechanism that detects novelty and initiates the creation of a new detector lies outside the present formalism and is described in the work by Parzhin et al. (2020) [1]. That work considers the architecture of a class of detector-neuron maps, where the central element governing its self-organization process is a "novelty" neuron.

## 15. Competition between Attractors.

The introduction of distance between attractors makes it possible to implement a Winner-Take-All (WTA) type competition process not only within a separate class neuron map, but also between maps of different classes. For this, it is necessary that the response of each detector-neuron contains summary information about the number of structural and parametric terms in the attractor structure. In our view, this approach may shed light on the well-known neural coding problem in neurobiology, specifically the Binding Problem [11] and the formation of invariant conceptual representations [23, 24], showing that the basic element of a neural code is, for example, not spike frequency or count, but the topology of coincidences in the dendritic tree [8] (see Section IV).

Let the set of attractors be finite and the distance between them be denoted as $d(\mathcal{G}_i, \mathcal{G}_j) > \delta$.

Strictly speaking, $d$ is a mixed pseudometric that combines discrete structures and parameters:

$$d(\mathcal{G}_1, \mathcal{G}_2) = \alpha \cdot d_{\text{struct}}(\mathcal{G}_1, \mathcal{G}_2) + \beta \cdot d_{\text{param}}(\mathcal{G}_1, \mathcal{G}_2) \qquad (114),$$

where:

- $\alpha$ is the weight of structural difference;

- $\beta$ is the weight of parametric difference.

The weight coefficients of structural difference can play a decisive role in determining the edit distance between attractors. The values of these weights are determined empirically.

Let us introduce the concept of *structural distance* [25]:

$$d_{\text{struct}}(\mathcal{G}_1, \mathcal{G}_2) = \min_{\mu \in \mathcal{M}}(\mid E_1 \setminus \mu(E_2) \mid + \mid E_2 \setminus \mu^{-1}(E_1) \mid) \qquad (115),$$

where:

- $\mu$ is a partial matching;
- $\mathcal{M}$ is the set of valid matchings.

$$\mu: E_2' \subseteq E_2 \to E_1 \qquad (116).$$

The term $E_1 \setminus \mu(E_2)$, where:

$$\mu(E_2) = \{ \mu(e_2) \mid e_2 \in \text{dom}(\mu) \} \qquad (117),$$

determines which elements of $E_1$ are not covered by the matching ("extra" elements in $\mathcal{G}_1$).

The term $E_2 \setminus \mu^{-1}(E_1)$, where:

$$\mu^{-1}(E_1) = \text{dom}(\mu) \qquad (118),$$

determines which elements are "extra" in $\mathcal{G}_2$:

$$E_2 \setminus \mu^{-1}(E_1) = E_2 \setminus \text{dom}(\mu) \qquad (119),$$

where $|\,\mathrm{dom}(\mu)\,|$ is the size of the matching structure.

Then

$$d_{\text{struct}} = |\,E_1\,| + |\,E_2\,| - 2 \cdot \text{max match} \tag{120}$$

Thus, $d_{\text{struct}}$ determines how many elements cannot be matched.

Let us introduce the concept of *parametric distance*:

After matching $\mu$:

$$d_{\text{param}} = \sum_{e \in \mathrm{dom}(\mu)} \sum_{i=1}^{k} \rho_i \left( m_i(e), m_i\big(\mu(e)\big) \right) \tag{121}$$

or via segments:

$$d_{\text{param}} = \sum_{e} \mathbf{1} \left[ Q\big(m(e)\big) \neq Q\left(m\big(\mu(e)\big)\right) \right] \tag{122}$$

Let's define the requirements for the Edit Distance Metric.

For $d$ to be a metric, it requires:

- Non-negativity: $d \geq 0$;
- Symmetry: $d(\mathcal{G}_1, \mathcal{G}_2) = d(\mathcal{G}_2, \mathcal{G}_1)$;
- Identity: $d = 0 \Leftrightarrow \mathcal{G}_1 \sim \mathcal{G}_2$.

Note that from a formal perspective, $d$ is still a *pseudometric*, since

$$\mathcal{G}_1 \sim \mathcal{G}_2 \neq \mathcal{G}_1 = \mathcal{G}_2 \tag{123}$$

Thus, we have formally defined a metric between attractors that enables the implementation of WTA-type competition.

It is worth noting that the proposed edit distance metric is not the only mechanism for realizing WTA competition. The choice of this mechanism within the current theory is justified by working with hypergraphs. Obviously, competition between biological neurons is based on other mechanisms, which we will discuss in Section IV. Moreover, as experimental results presented in Section III show, the proposed approach is overly simplified and contains a number of critical drawbacks. Overall, the problem of defining a metric for WTA competition remains open.

The described ISL process (step by step), as well as the stages of inference and self-organization of detector-neuron maps, are presented in Figure 1.

## 16. Discussion of Theoretical Results.

The presented theory is conceptual in nature and requires further research. Below, we will focus on several debatable questions that arose in the process of formulating the theory:

1. The work is devoted to self-organizing systems, yet primitive detectors are given exogenously rather than formed during the learning process. This is a model limitation, but in the brain, the formation of these areas - directly connected to the sensory system - is likely genetically predetermined rather than formed through learning. However, the problem of the necessary and sufficient set of primitives to form all possible attractors remains pressing. In the experimental part of the work, we minimized the number of extracted primitives as much as possible, approximating the image contour with straight-line segments. This led to a loss of important information about curves in the image contour and introduced numerous additional errors and distortions during the approximation and skeletonization of contours. This approach significantly reduced classification accuracy.

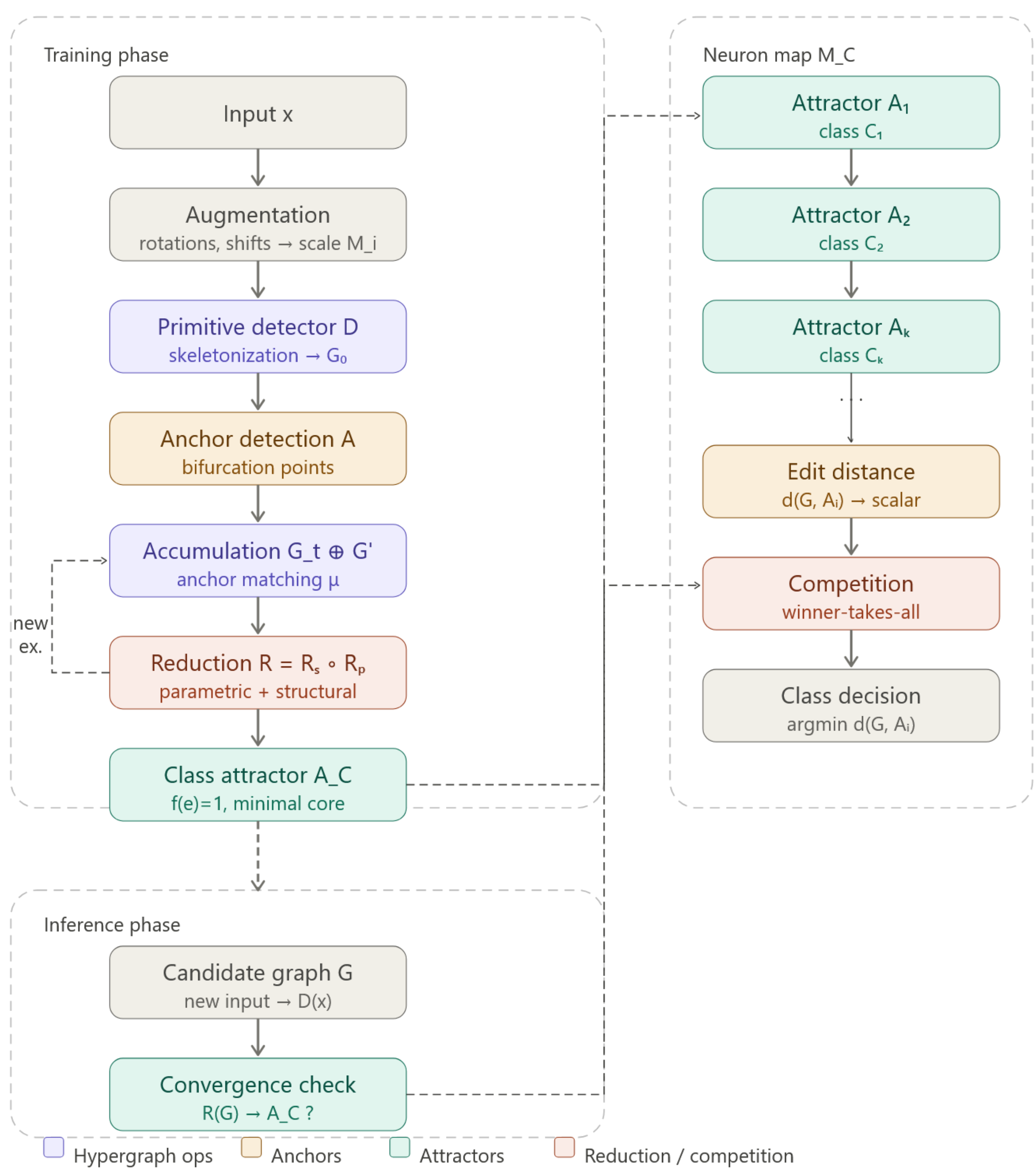

Figure 1. ISL concept diagram.

2. In the experimental part of the work, we also exogenously specify coordinate systems and measurement scales. The problem of variations is a broader issue associated with primitive feature detection. It is likely that endogenous scales and coordinate systems are formed in the brain based on both genetics and learning. This problem is also a separate direction and lies beyond the scope of our research.

3. One of the central problems of this theory is finding the critical structural (anchor) points that form the basis of the structural attractor. The presented theory demonstrates that these points are determined during the learning process as a result of applying the reduction operator, based on the frequency of occurrence of candidate points in the hypergraph representation of compared images. However, in most theoretical discussions, we assume that the initial hypergraph structure has already been constructed and candidate anchor points (elements) have already been defined as the most informative elements.

This process of searching for anchor point candidates aligns with the well-known neurobiological principle of *Good Continuation*. In gestalt psychology, Good Continuation means the tendency to perceive elements arranged along a smooth spatial trajectory as parts of a single continuous structure. A quantitative study by Field, Hayes, and Hess (1993) showed that local orientation alignment of neighboring elements facilitates their integration into a single contour [26]. However, this principle may reflect the process of local parametric reduction even at the stage of hypergraph construction (prior to building the accumulated configuration), which isolates candidate anchor points within the hypergraph structure.

*Clarification:* Even at the stage of constructing the initial hypergraphs of any incoming images, we use the same parametric reduction operators as in the attractor formation process, but solely for the local isolation of parametrically invariant substructures without applying structural reduction. It is this parametric reduction that allows us to build the hypergraph "superstructure" over the base spatial graph, isolate parametrically stable substructures, and identify anchor candidates. That is, we form the following transformation chain:

"Input object → Primitive extraction → Local parametric reduction (structural invariant detection) → Hypergraph construction → Global structural reduction → Class attractor".

However, the mathematical formalization of local parametric reduction is an independent task and lies beyond the scope of this work.

4. An important problem is defining the invariant parameters of the "global" topological connectivity of the structure, which prevent the structure from "spreading out" during the reduction and subsequent "gluing" process. In this theory, such parameters include: closedness or disconnectedness of the structure, preservation or violation of monotonicity in parameter changes of structural elements, contour traversal direction, etc. This process defines yet another level of attractor dynamics and requires deeper theoretical consideration. However, note that in biological systems, where—in our view—the spatial topology of the stimulus does not necessarily have to be preserved in the structure of the dendritic tree, this problem of global invariants looks different (see Section IV).

5. One of the most challenging problems is modeling the attention operator. Modeling the attention system is based on neurobiological research of active perception and, in our view, requires further formalization as one of the fundamental processes in solving the NP-hardness problem. In this work, we used only a heuristic approach to solve this problem.

6. One of the advantages of this approach is the formation of the system's natural endogenous ontology—its internal model of the "external world." This ontology is formed not only by individual detector-neurons or class neural maps, but also by their hierarchy, i.e., *the hierarchy of attractors* [1].

7. The scaling potential of structural feature-based models is demonstrated in works [5, 6]. This suggests that the proposed approach may also possess a similar and possibly even greater scaling potential. Indeed, when constructing, say, a hypergraph of a 3D image, the anchors can be segments or arcs rather than points, and planes or curved surfaces can be used instead of the segments forming the spatial structure in the 2D model. We also address another scaling problem related to system overfitting: detector-neurons are trained independently, and it is their hierarchy that creates *the hierarchy of concepts*

*(attractors)*. The proposed approach has neurobiological justification [24] and separates the processes of attractor formation for individual simple images (e.g., closed contours), complex images (scenes), movements, processes, and semantic constructs across different hierarchy levels. This opens up new directions for future research.

8. A key idea of this approach is that we view the neuron's dendritic tree as the primary "computational" system that forms the structural attractor, while the neuron's response reflects its generalized characteristic (such as its complexity). This hypothesis agrees with neurobiological studies showing that individual dendritic branches function as independent computational units, and neuronal excitation can be determined by the activity of a restricted subset of synapses [15, 27, 28].

9. One of the most significant problems of the proposed mathematical theory is hypergraph matching, which is generally an NP-hard problem. We attempt to alleviate this difficulty by considering relatively simple MNIST structures and modeling active perception. However, as stimulus structures grow more complex (e.g., when considering complex three-dimensional images), this problem may become critical. Furthermore, it is unlikely that brain neurons perform hypergraph matching or compute their edit distances. In Section IV, we offer a radical perspective on solving these problems in biological neurons and neuromorphic systems. The proposed ISL theory should be viewed not as a mathematical model of low-level physical brain simulation, but as a mathematical abstraction of information processes occurring at the neuronal level. It answers the questions "what is being done?" and "why?".

10. In the theoretical part of the work, we singled out *associative recognition* as a separate research stage. If recognition is interpreted as the dynamical relaxation of the system to a structural attractor, reaching this attractor does not necessarily have to terminate the computational process. On the contrary, the formed attractor can serve as a source for the further spread of internal activity across its own structure, helping to restore missing elements of the object's internal representation. Such a mechanism can underlie associative recognition and provide a computational interpretation for a number of top-down perception modulation effects. In the experimental part, we also do not consider associative recognition: if the initial image contour contains unwanted gaps or incomplete structure, we do not include them in the training set. If these gaps or losses of structural elements occur during preprocessing (binarization, skeletonization), they are not associatively restored, which generally can reduce classification accuracy. Given this circumstance, we also exclude images with initially incomplete or broken contours from the test set.

11. To define the position of our theory within the ML landscape, we note that ISL does not belong to the statistical paradigm of Statistical Learning Theory (Vapnik, Valiant), because learning is not formulated as the minimization of expected risk, estimation of a probability distribution, or optimization of a loss function. Conceptually, ISL is closer to the idea of Concept Learning and Version Space (Tom M. Mitchell (1977)) [29], where learning is understood as the sequential refinement of internal representation based on structural constraints. However, unlike classical Version Space, ISL does not assume a fixed hypothesis space: the representation architecture itself changes (self-organizes) during the learning process through the formation of structural attractors.

**17. Conclusions of the Section.**

In the first part of this work, a formal theory of learning called ISL (Invariant Structural Learning) was proposed, viewing learning as a process of structural self-organization rather than a statistical optimization task. Unlike traditional machine learning methods based on minimizing an externally defined loss function, ISL formulates learning as *the convergence of a sequence of structural transformations to a fixed point of the reduction operator on the space of hypergraphs*.

The main theoretical result is the introduction and mathematical formalization of the *structural attractor* as a stable representation of a concept. The work proves that the structural attractor is a well-defined mathematical object within the proposed model and possesses several fundamental properties. In particular:

- The structural reduction process terminates in a finite number of steps bounded by the initial structural complexity of the object (Theorem 1);
- Under the conditions of anchor connectivity, the structural attractor exists, is unique up to structural equivalence, and is independent of the order in which training examples are presented (Theorem 2);
- The set of attractors self-organizes into a map of distinguishable fixed points without a pre-specified number of subclasses (Theorem 3).

These results allow us to view a concept not as a set of statistical parameters, but as a stable structure of invariants arising from the structural-parametric reduction process. At the same time, the reduction process can form not just a single realization of the attractor, but a parametric family of permissible representations of a single class, united by a common structure of invariants (Theorem 3) and forming an ontological structure of concepts.

It is shown that the structure of an observed object is naturally described by a two-level model. The first level forms the base graph, determining the topology of interrelationships between elements. The second level represents the parameter hypergraph formed by coordinate systems and measurement scales. Such a separation makes it independently possible to formalize the extraction of structural and parametric invariants, thereby dividing the processes of forming the concept's structure and its parametric variability.

Unlike classical data augmentation, which increases the volume of the training sample, the proposed approach views augmentation as a mechanism for constructing the class's internal parametric metric. The sequence "structure → segmentation → metric" determines the internal geometry of the space of permissible concept realizations and explains the possibility of learning from an ultra-small number of initial examples.

A fundamental difference between ISL and modern ANNs lies in the nature of the formed fixed point. The weight matrix of a trained neural network can be viewed as a fixed point of a loss function minimization operator, but such a fixed point is not an attractor in the ISL sense. In the first case, stability is determined by the geometry of an externally defined error function common to all classes simultaneously. In the second case, stability is determined exclusively by the internal structure of the class itself and arises without using an external optimality criterion. Thus, learning acquires an endogenous character and is determined entirely by the structure of the observed data.

The proposed model possesses a number of core properties that distinguish it from existing statistical learning approaches:

- Absence of the need to define a loss function and an external optimization criterion;
- Absence of backpropagation as a learning mechanism;
- Learning in a single pass over the training examples;
- Ability to learn under conditions of ultra-small training sets (few-shot learning);
- Learning exclusively from positive examples of each class without constructing separating hyperplanes;
- Inherent (not post-hoc) interpretability of decisions made thanks to the explicit representation of the concept as a structural attractor.

Thus, the theoretical part of the work formalizes an alternative non-optimization learning model based on structural self-organization and invariant extraction. Within the framework of the proposed approach, information is viewed not as a statistical characteristic of a set of observations, but as a stable structure emerging during the reduction process and embodied in the structural attractor, creating a mathematical foundation for subsequent investigation of potential neurobiological mechanisms of concept formation and the origin of neural coding.

## III. Experimental Verification on MNIST

### 1. Goals and Limitations.

This section presents an experimental implementation designed to verify the properties derived from the proposed theory:

1. Existence of a structural attractor - Theorem 1;
2. Uniqueness of the structural attractor - Theorem 2;
3. Conditional uniqueness of the class structural attractor - Corollary 1;
4. Convergence to the attractor in a finite number of steps - as a consequence of Theorem 1;
5. Self-organization of the class detector-neuron map (the "alphabet" of concepts) - Theorem 3;
6. Sufficiency of an ultra-small sample for training without backpropagation and in a single pass (one training epoch) - Corollaries 3, 4;
7. The role of augmentation as an operator exploring the parametric space of the structure - Corollary 2;
8. Invariance of the structural attractor to the order of presentation of examples during the learning process (accumulation order invariance).

We did not set out to achieve SOTA results on classical benchmarks. Our goal is to empirically prove the viability of the proposed theory at the PoC level.

To test the theoretical propositions, we construct a concept-attractor classifier and evaluate it on a subset of MNIST with clean, unbroken contours to demonstrate that it simultaneously achieves high accuracy under ultra-small training sample sizes (few-shot learning) and local explainability of decisions made. To solve this problem, we developed the *Compartmental Attractor Neural Network* (ComAN), available on GitHub (link at the end of the work).

The choice of the MNIST benchmark as an experimental base is justified by the following considerations:

1. We examine a biologically grounded image processing chain: *"sensory stimulus perception → primitive extraction at the preprocessing stage → holistic image attractor construction during learning, or its activation during inference → decision-making on stimulus class membership via WTA competition of activated attractors during inference".*

However, the goal of this work is specifically to investigate the formation of the structural attractor; therefore, the stage of extracting primitives from "raw" images is not examined in detail (in terms of biological plausibility) and is modeled using established procedures and algorithms. That is, we do not work with "raw" pixel images at the input of our model, unlike classical ANNs. This makes it difficult to use more complex benchmarks like ImageNet, CIFAR, or even Fashion-MNIST for primitive extraction. Furthermore, at this stage, we demonstrate the formation of lowest-level structural attractors for the simplest two-dimensional contour structures that lack background or individual substructures forming a more complex scene. Such an approach has neurobiological backing (David Marr, 1982) and is the subject of modern research by Gemma Taylor et al. (2014) [30];

2. Our task is to obtain a minimal vocabulary of detectable primitives and their parameters to simplify the construction, matching, and reduction of hypergraph structures. The experiment is intentionally restricted to images with clean digit contours, which corresponds to the neurobiologically grounded baseline recognition mode based on complete structural descriptions.

Images with broken or fragmented contours are excluded from this study because their classification requires an additional *associative recognition mechanism*, which calls for a different mathematical approach and has a different neurobiological motivation [31].

In the most general sense, the construction of structural attractors can be represented as mapping observed objects (digits of the same class) into hypergraphs, which during training are transformed into a cumulative hypergraph followed by its reduction to an attractor. One of the most complex and computationally expensive procedures during training and inference is hypergraph alignment for matching, which is generally an NP-hard problem. This procedure maps elements of different hypergraphs into a common structural space, forming a cumulative hypergraph where element occurrence frequencies and their stabilization under reduction can be analyzed.

Note that the theoretical foundation of this study is formulated in the formalism of hypergraphs with parametric hyperedges linking sets of parameters across various vertices. However, in the experimental implementation, the parametric layer is reduced to a group of attributes for each vertex rather than an explicit hypergraph edge: each vertex carries a vector of thirteen continuous and categorical attributes (Table 1) rather than participating in a separately stored relation. Thus, the base graph is a directed attributed graph in NetworkX, and the hyperedge layer appears implicitly via a diagnostic-weighted node substitution cost that links attributes between matching vertex pairs during comparison. This is a model simplification rather than a theoretical departure: hypergraph terminology is retained for the general formalism, while the term "graph" is used in the experimental description sections where this simplification is valid.

## 2. From Raster to Graph: Implementation of the Primitive Space.

The conversion of a raster image into an attributed graph is implemented as a three-stage pipeline:

1. *Binarization:* The image is binarized using a fixed intensity threshold that separates foreground and background pixels.
2. *Skeletonization (Growing Neural Gas):* The Growing Neural Gas algorithm [32] adaptively places vertices along the object's midline, creating a topologically accurate skeleton that accounts for the local distribution of foreground pixels.
3. *Polygon Simplification (Ramer-Douglas-Peucker):* The Ramer-Douglas-Peucker algorithm [33] removes intermediate vertices whose deviation from the chord between their neighbors is less than a permissible value, retaining only structurally significant points: endpoints, corner points, and junction (bifurcation) points.

As a result, a bipartite directed attributed graph is produced.

- *Vertices of the first kind (Point nodes)* represent critical points of the skeleton and correspond to Definition 1.

– *Vertices of the second kind (Vector nodes)* represent directed segments between adjacent Point nodes and correspond to structural edges after their conversion into first-class objects.

Thus, each segment can carry its own group of parameters, as required by the theoretical propositions on the stratification of the parametric space over the structure. The spatial coordinates of all vertices are normalized relative to the center of mass of the foreground mask into a symmetric range, which ensures invariance to translation in the image plane.

Each vertex contains a *thirteen-element feature vector*. The features are divided into four conceptual categories:

– *Spatio-geometric:* normalized_x, normalized_y (horizontal and vertical coordinates relative to the centroid); distance_to_centroid (radial distance).
– *Orientational* (defined only for Vector nodes): horizontal_direction, vertical_direction (categorical orientation along two coordinate axes); angle_with_ox (orientation angle in degrees, also defined for Point nodes as the convergence angle of adjacent segments).
– *Structural-functional* (defined only for Point nodes): is_endpoint, is_corner (boolean flags indicating whether a node terminates a segment or registers a directional change exceeding a fixed threshold, respectively); junction_angle_min (smallest internal angle at a bifurcation point).
– *Topological neighborhood:* length_ratio_to_max (segment length normalized by the longest segment in the graph); eccentricity (graph-theoretic node eccentricity, i.e., maximum distance to any other node); mean_neighbor_vector_length (average normalized length of adjacent segments); neighbor endpoint count and neighbor junction count (number of adjacent Point-type nodes by role degrees).

The complete set is given in Table 1. Features defined exclusively for one node type - for example, the orientational triplet for Vector nodes or the structural-functional triplet for Point nodes - are skipped by default during comparison when matching against a node of a different type. Thus, the cost decomposition (discussed in detail in [4]) always operates on the intersection of features actually present on both sides of the proposed substitution.

Table 1. The thirteen-element node feature vector of the graph image representation, grouped by conceptual category.

| Name | Description |
| --- | --- |
| Normalized x-coordinate | Horizontal coordinate of the node normalized against the centre of mass of the foreground mask; locates the critical point of the skeleton along the horizontal axis of the image plane. |
| Normalized y-coordinate | Vertical coordinate of the node normalized against the center of mass; locates the critical point along the vertical axis. |
| Distance to centroid | Normalized radial distance of the node from the center of mass; distinguishes central skeletal points from peripheral ones. |
| Angle with OX | Orientation angle of the segment in degrees for Vector nodes; angle of convergence of incident segments for Point nodes; an invariant shape descriptor of the segment. |
| Endpoint flag | Boolean feature; affirmative when the node has degree one and therefore terminates a single segment of the skeleton. |
| Corner flag | Boolean feature; affirmative when the node registers a direction change exceeding a fixed angular threshold. |
| Junction angle (min) | The smallest internal angle at a bifurcation point between converging segments; distinguishes acute from open junctions of the skeleton. |
| Neighbor endpoint count | Number of adjacent Point nodes flagged as endpoints; characterizes the local density of dead-end branches near the node. |
| Neighbor junction count | Number of adjacent Point nodes flagged as junctions; characterizes proximity to complex structural articulations. |
| Average neighbor vector length | Mean normalized length of segments incident to a Point node; characterizes the local geometric scale around the point. |
| Horizontal direction | Categorical feature defined for Vector nodes; orientation of the segment along the OX axis (left or right). |
| Vertical direction | Categorical feature defined for Vector nodes; orientation of the segment along the OY axis (up or down). |
| Length ratio to maximum | Ratio of a Vector node's length to the longest segment in the graph; distinguishes principal strokes from short connecting segments. |

**3. Reduction Operator and Attractor Formation.**

Let us show how the abstract reduction operator $R$ is implemented iteratively, and demonstrate the convergence of the reduction process using the formation of attractor 7_1 (the sole attractor of class 7 in the alphabet) as an example.

To form the attractor during the learning process, we manually selected nine original images of the digit "7" from the MNIST set, skeletonized them using the Growing Neural Gas method, and simplified them using the Ramer-Douglas-Peucker algorithm. The general pipeline of the classification process using the digit "7" as an example is presented in Figure 2.

The condition for manual example selection is merely the requirement of:

- *Contour integrity:* The absence of extraneous gaps in the original image contour;
- *Structural completeness:* The presence in the contour of all elements necessary for recognition (the hidden attractor).

Due to variations in handwriting, example graphs differ by the presence of intermediate corner points along the horizontal stroke, slight bends in the diagonal stroke, and variations in the size of the upper angle. For any example in any recognition class, the same graph edit distance algorithm is used, which for each example finds an invariant core containing critical structural elements - anchors. The search for and isolation of such anchor elements (implementing the attention operator $\mathcal{F}$ (100, 102)) models the neuropsychological process of active perception - the shifting of the focus of attention (visual focus) - and is an important step in solving the graph matching problem.

For the digit "7," this core consists of three critical points (Figure 3):

- The starting endpoint of the horizontal stroke in the upper-left corner of the field - the contour "capture" point (the start point of the contour traversal during the sequential formation of the graph structure);
- The upper corner point in the upper-right corner (with a non-monotonically changing segment orientation parameter during sequential contour traversal), where the horizontal stroke transitions into the diagonal;

– The endpoint of the diagonal (segment) in the lower part of the field.

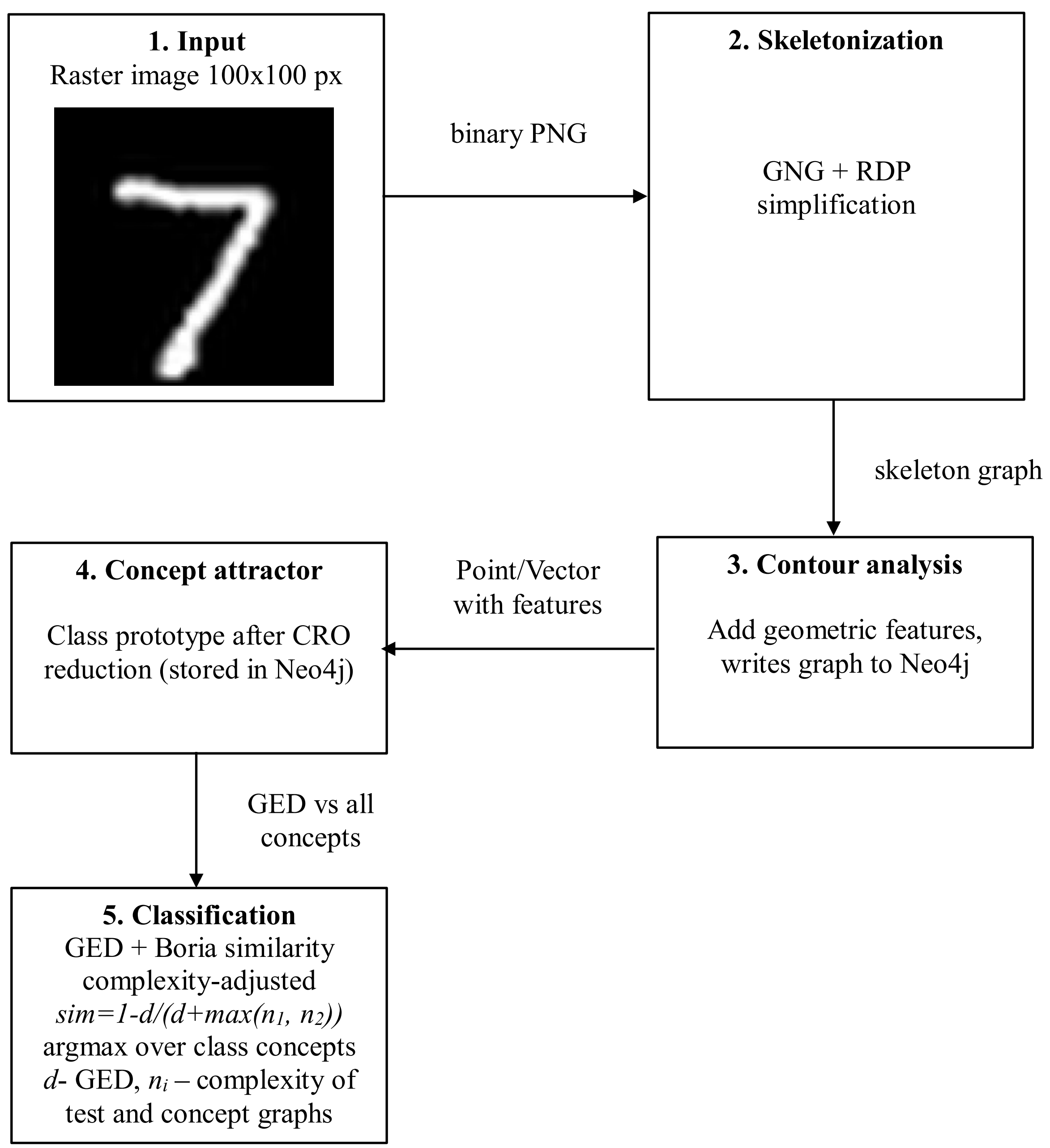

Figure 2. End-to-end pipeline from raster image to concept-attractor classification via graph edit distance.

These three Point-type vertices are connected by edges to two Vector-type vertices: one for the horizontal segment between the starting point and the upper corner point, and one for the diagonal segment between the upper corner point and the endpoint. All other vertices appearing only in a subset of the original elements - additional intermediate corners on the horizontal stroke, micro-bends along the diagonal - are removed in the next step as elements with a frequency of $f(e) < 1$.

The parameters of these vertices (only coordinate parameters are shown in the figure for illustration) lie within the $[-1, +1]$ range and indicate the *minimum/maximum* limit values.

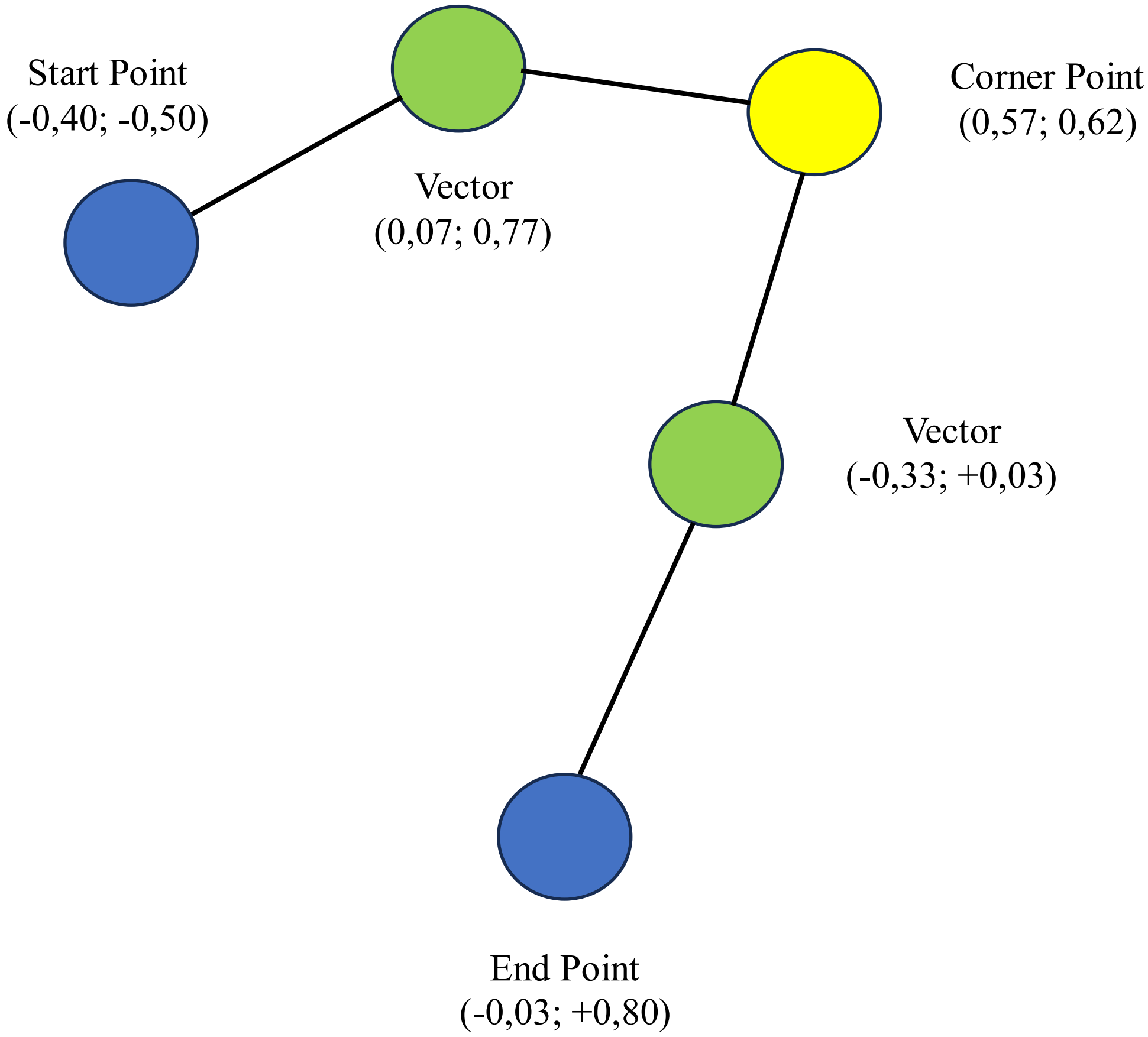

Figure 3. Concept-attractor 7_1 after structural reduction: a linear chain of three Point and two Vector nodes - one of the smallest nontrivial attractors in the alphabet.

Attractor 7_1 in Figure 3 illustrates what the fixed point of the reduction operator looks like.

For each preserved vertex of the invariant core, the reduction operator records the interval of permissible values for each coordinate characteristic across all absorbed source data. The intervals reflect the expected topology of the digit: the starting point is localized in the upper-left corner, the upper corner point in the upper-right, and the endpoint in the lower half with a relatively wide horizontal latitude. Vector nodes inherit similar intervals: the horizontal segment is stably positioned in the upper band with a small spread, and the diagonal segment crosses the field from the upper-right region to the lower band. The width of the resulting interval directly determines the diagnostic weight: coordinates (parameters) whose values repeat stably across all source data have a greater weight during classification.

After several training iterations, the structure stabilizes into a linear bipartite chain alternating between Point and Vector vertices: starting point → horizontal vector → upper corner point → diagonal vector → endpoint. Subsequent iterations may expand the intervals but do not introduce new vertices. This is a sign of reaching the attractor (the stopping criterion).

The resulting compact attractor of the digit "7" has 3 Point-type vertices and 2 Vector-type vertices connected by 4 edges, and a complexity $C$ (see simplified expression (60)) equal to 9, making it one of the smallest nontrivial attractors in the alphabet. The compact structure encodes the topological invariant of the digit "7", but this same compactness is a source of excessive activity - a small, broadly segmented attractor is constantly activated and competes with more complex concepts.

Thus, we have examined the three main steps corresponding to the operation of the generalized reduction operator (26):

- ***Step 1*** **-** *Parametric Reduction* ($R_p$). For each vertex present in the current concept and in the new example, continuous attributes are updated: the interval of permissible values is expanded to the extreme observed values, and the typical value is fixed at the center. Categorical attributes are formed as the reduced intersection of permissible values.

- ***Step 2 -*** *Structural Reduction* $(R_s)$. Vertices that do not appear in all absorbed examples $(f(e) < 1)$ are removed (in this example, intermediate corner points of the horizontal stroke were present only in some of the originals and were removed). Five vertices of the invariant core remain.
- ***Step 3 -*** *Connectivity Restoration of the Resulting Graph* $\mathcal{G}$ (the Closure operator (19)). The remaining vertices are connected into a single graph - the attractor - while preserving the overall spatial topology.

The reduction iteration stops when the next absorbed example does not change the attractor structure. This is the signal that the fixed point of operator $R$ - the attractor according to Theorem 1 - has been reached.

**4. The Role of Augmentation: Parametric Stratification.**

The experiment clearly demonstrates that augmentation builds a metric via segmentation rather than increasing the sample size in the classical sense.

During augmentation, each original example is expanded into a family of variants with an equivalent structure and independently varying parameters. Variants are created using a parametric augmentation procedure, where each original is randomly rotated by an angle in the range $[-10°, +10°]$ and shifted by an amount in the range $[-10\%, +10\%]$ of a $100 \times 100$ pixel window size. The range boundaries are chosen empirically. The number of variants per original is also determined empirically and ranges from five to ten, depending on the parametric variability of the examples.

Complete data on the number of examples in the training set, as well as the number of anchor points and spatial connections between them in the concept structure, are presented in Table 2.

The experiment confirms the following theoretical results:

1. The augmented family reinforces the bounds of permissible coordinate intervals at each anchor and narrows the estimation of diagnostic weight. Without augmentation, an interval estimated from only three to nine source data instances converges to an almost degenerate width across most coordinates, and the diagnostic weight saturates uniformly across all features (parameters).

2. The segmentation $Q$ of each parameter is defined by a stable interval rather than a single point. This directly implements the theoretical principle of metric-from-segmentation: "structure → segmentation → metric". The interval width after reduction governs the contribution of each feature to the substitution cost, with narrower widths receiving a higher weight during comparison.
3. The augmented family empirically ensures that learning is invariant with respect to order: after merging original structures in any order with their variants, the resulting attractor is invariant to the permutation of the merge order, since the set of elements with $f(e) = 1$ on the cumulative graph does not depend on the order in which training examples are presented.

Thus, the purpose of data augmentation is not to increase the training sample size, but to form an internal metric based on the input data.

Table 2. Concept alphabet: per-concept origin counts, augmented-variant counts, post-augmentation training-set size, concept node count, and class membership.

| **Concept** | **Class** | **Originals** | **Augmented** | **Training instances** | **Concept node count** |
|---|---|---|---|---|---|
| 0_1 | 0 | 3 | 30 | 33 | 10 |
| 1_1 | 1 | 3 | 30 | 33 | 7 |
| 1_2 | 1 | 5 | 50 | 55 | 3 |
| 2_1 | 2 | 11 | 110 | 121 | 7 |
| 2_2 | 2 | 5 | 50 | 55 | 12 |
| 3_1 | 3 | 7 | 70 | 77 | 9 |
| 4_1 | 4 | 6 | 30 | 36 | 8 |
| 4_2 | 4 | 6 | 59 | 65 | 9 |
| 5_1 | 5 | 8 | 80 | 88 | 7 |
| 6_1 | 6 | 7 | 70 | 77 | 10 |
| 7_1 | 7 | 9 | 90 | 99 | 5 |
| 8_1 | 8 | 7 | 35 | 42 | 11 |
| 9_1 | 9 | 8 | 80 | 88 | 8 |
| Total - 13 | 10 | 85 | 784 | 869 | — |

**5. Classifier Operation Principle.**

Classification (inference) of unknown (test) images boils down to computing the upper bound of the graph edit distance between the initial skeleton graph of the test image and each concept-attractor in the alphabet - practically speaking, to computing the cost of reducing the input graph to the hypothesized attractor (if such a reduction is even possible). To do this, we introduce *edit operation costs*, where the cost of vertex substitution is determined by how much its attributes deviate from the attractor's attributes, with a higher weight assigned to diagnostically stable features [4].

The implementation utilizes an iterative upper bound generator `optimize_graph_edit_distance` from NetworkX [34], which is initialized by polynomial approximation based on bipartite assignments [35] and monotonically refines the upper bound by iterating over alternative vertex matchings. After an empirically set fixed time budget - 5 seconds per concept comparison - the procedure stops and returns the best obtained upper bound. The returned value is an estimate over a limited time frame: it does not guarantee finding the global minimum of the edit distance, but on concept-attractor graphs with sizes and structures ranging from 3 to 12 reference nodes (Table 2), the estimate stabilizes near the optimum in more than 99% of comparisons. This approach, along with modeling the attention operator $\mathcal{F}$ (100), demonstrates an attempt to mitigate the NP-hard problem of graph matching and alignment.

Edge processing is also intentionally simplified: any pair of edges is considered compatible, edge deletion costs MINOR, edge insertion is free, and no explicit attribute substitution is computed at the edge level. Therefore, the entire structural penalty is concentrated at the vertex level via a cost scale and diagnostic weighting (the empirical cost scale and cost calculation are discussed in [4]).

To predict the class, we implement a Winner-Take-All (WTA) competition principle among simultaneously excited (activated) detector-neurons based on comparing their responses, which generally represent the complexity values of their concept-attractors. However, this complexity is determined not simply by summing the number of structural elements in the attractor and the connections between them, but also by their parameters - the ranges of the 13 features on each node (coordinates, angles, direction, cycles, etc.) and their diagnostic weights $1/(\text{range width} + \varepsilon)$ [4]. This allows for a more accurate implementation of the WTA competition process, where the winner is the detector-neuron with the highest calculated attractor complexity value.

Furthermore, along with the predicted class, the classifier returns an *explanation artifact* containing fields such as: Winner concept identifier; Likelihood function value; Complexity function values.

These fields provide a complete local explanation of the decision: any subsequent verification of the prediction reduces to analyzing why the winning concept ranked higher than its closest competitor, without needing any post-hoc model.

**6. Results of Experimental Studies.**

The training set consisted of 85 original MNIST digits across 10 classes (Table 2), featuring a complete hidden attractor distributed across 13 concepts (3 to 11 originals per concept), augmented up to 869 instances. The condition for selecting originals is solely the presence of a complete, unbroken contour that allows for a subjectively unambiguous classification of the image. Three additional concepts (subclasses) for the digits "1", "2", and "4" were formed empirically based on significant structural differences in their writing styles.

Table 2 shows that attractor complexity cannot be determined solely on the basis of a simple count of graph vertices and edges, because for many concepts - such as 1_1 and 5_1, 4_1 and 9_1, 3_1 and 4_2, 0_1 and 6_1 - these counts coincide. This indicates that a clear separation of attractors of different classes by their complexity must occur at the parametric level. This conclusion calls for a closer focus on developing a mechanism to account for the influence of the parametric layer on the overall complexity of attractors.

Test images were not used in concept formation. Out of 8,707 MNIST test images chosen randomly from the general library, 8,685 (99.75%) yielded a valid skeleton graph;

22 images fell into a dead-letter queue due to a disconnected graph during binarization and skeletonization and were excluded from metric calculations as preprocessing errors. On the valid set, the following results were obtained: *accuracy 91.13%, weighted precision 91.92%, recall 91.13%, F1-score 91.34%* (Table 3). Forty-four images (0.5% of the valid set) were not assigned to any concept; these were concentrated in classes "6" (16 images) and "8" (14 images).

Table 3. Per-class classifier metrics on the complete-contour subset of MNIST: precision, recall, F1, support.

| Class | Precision, % | Recall, % | F1, % | Support |
|---|---|---|---|---|
| 0 | 99.9 | 91.9 | 95.7 | 830 |
| 1 | 93.7 | 96.5 | 95.1 | 1122 |
| 2 | 92.3 | 77.6 | 84.3 | 955 |
| 3 | 91.8 | 95.8 | 93.7 | 968 |
| 4 | 92.1 | 84.0 | 87.9 | 950 |
| 5 | 94.0 | 92.0 | 93.0 | 796 |
| 6 | 84.9 | 92.4 | 88.5 | 741 |
| 7 | 82.0 | 95.2 | 88.1 | 1011 |
| 8 | 98.7 | 94.5 | 96.6 | 581 |
| 9 | 92.8 | 92.3 | 92.6 | 731 |
| **Total** | **91.92** | **91.13** | **91.34** | **8685** |

The 0.79 percentage point difference between weighted precision and recall indicates a generally uniform coverage of classes by the attractor alphabet. The residual unevenness is concentrated primarily in class "2" (with a recall of 77.6%) and class "4" (with a recall of 84.0%). Both classes are characterized by high precision and lower recall, which may indicate insufficient coverage of the class by the attractor alphabet.

The complete confusion matrix is presented in Figure 4.

Table 4 demonstrates the ten largest mismatch pairs by absolute count. Let us consider the dominant pattern: the "2" → "7" mismatch (111 cases, 51.6% of class "2" errors), as well as the "4" → "7" (62 cases), "1" → "7" (26 cases), and "3" → "7" (10 cases) patterns, which share a single structural cause.

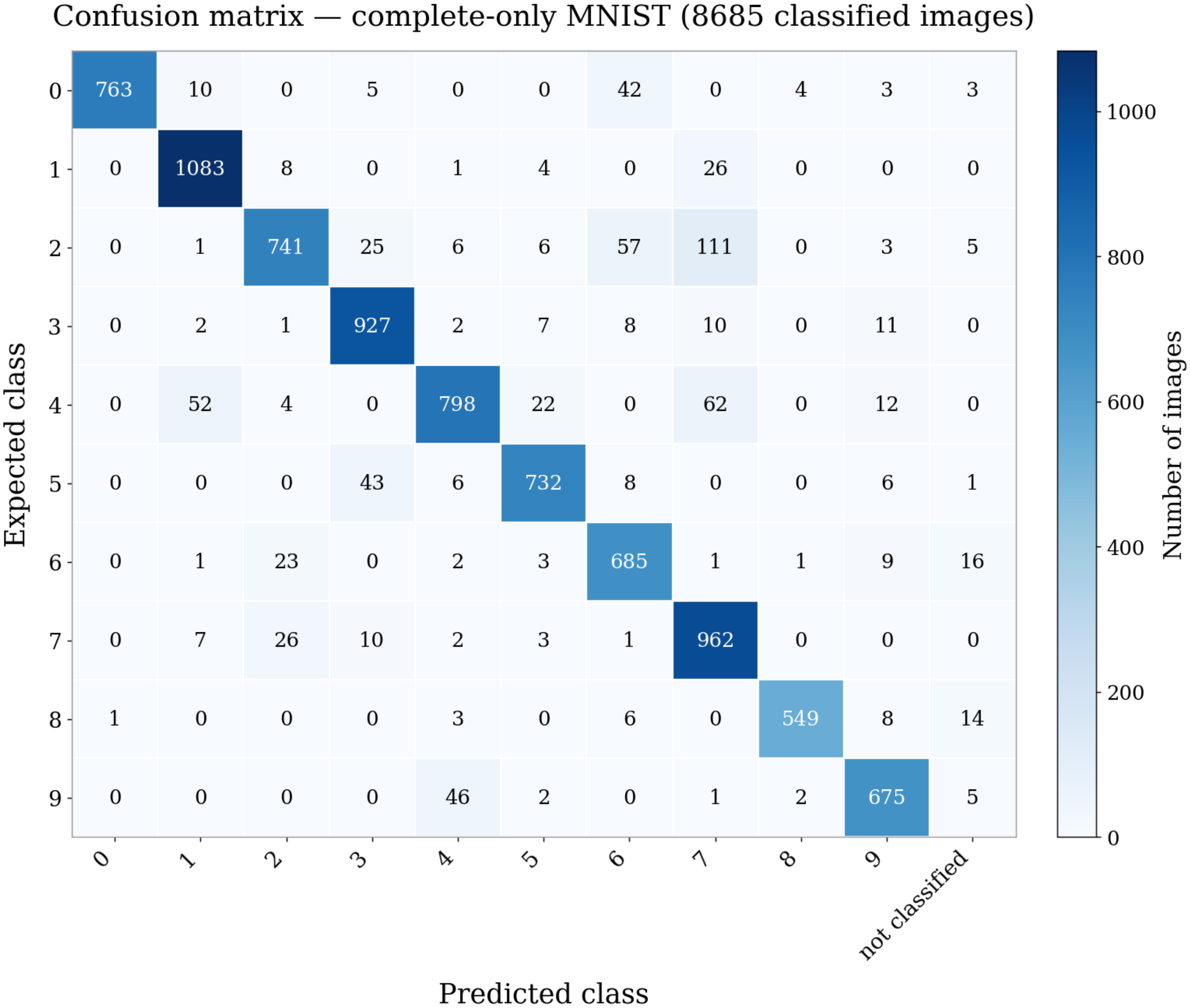

Figure 4. Confusion matrix on the complete-contour subset of MNIST (8 685 successfully classified images across 10 classes). The 'not classified' column contains images that did not match any concept.

The compact attractor 7_1 with five Point + Vector vertices encodes a linear topology of two segments and one corner inflection. This topology is locally consistent with stylistic variants of digits: "2" with an open tail (an angle in the upper-left part, next to an endpoint with an open loop and a diagonal segment down to the lower endpoint);

elongated forms of the digit "3" without a closing arc; segments of the digit "4"; and variants of the digit "1" with a slanted upper stroke.

Table 4. Top-10 confusion pairs (true → predicted): absolute count and share of class-level errors.

| True class | Predicted class | Count | % of class errors |
|---|---|---|---|
| 2 | 7 | 111 | 51.6 |
| 4 | 7 | 62 | 40.0 |
| 2 | 6 | 57 | 26.5 |
| 4 | 1 | 52 | 33.5 |
| 9 | 4 | 46 | 75.4 |
| 5 | 3 | 43 | 67.2 |
| 0 | 6 | 42 | 62.7 |
| 7 | 2 | 26 | 50.0 |
| 1 | 7 | 26 | 56.5 |
| 2 | 3 | 25 | 11.6 |

The comparator modeling the WTA competition process in many of these cases favors attractor 7_1 based on considering the diagnostic weighting of node substitution cost [4]. Let us examine why this happens using the example of attractor 7_1 winning when presented with the digit "2".

Each attractor node stores an interval $[\min; \max]$ for each of the 13 features; the feature weight $w = 1/(\text{interval width} + \varepsilon = 0.1)$; values falling outside the interval incur a full penalty $\text{NO_MATCH} = 1.0$. A narrow interval defines a high weight, meaning the node is more "demanding" of the feature value, whereas a wide interval provides "tolerance" for a specific feature value. For attractor 7_1, all 5 nodes substituted within their intervals (costs $0.04 - 0.39$, similarity $0.8317$), whereas for attractor 2_1, three features fell outside the specified intervals (distance_to_centroid $0.558 \notin [0.69; 1.00]$ at the starting point, normalized_x $0.200 \notin [-0.70; 0.10]$ at the corner point, length_ratio_to_max $0.306 \notin [0.43; 1.00]$ at the horizontal vector). This led to each "dropped" feature incurring a full penalty - resulting in a similarity of $0.7300$. That is, in these examples, none of the considered classes has an attractor whose node substitution cost on such impoverished graphs would be lower than that of attractor 7_1: competitors have some features falling outside the learned intervals, whereas 7_1's tolerances accept this configuration without penalties.

This indicates the imperfection of the method for computing the attractor complexity measure based on the applied graph edit distance calculation methodology [4] and the empirical approach to defining the ranges of varying parameters. That is, from the perspective of ISL theory, in these cases, the projections (complexity values) of the full invariant signatures (66) of different attractors in the responses of these detector-neurons are indistinguishable.

For some of the considered examples, the reason for the "victory" of the simpler attractor 7_1 is that more complex attractors do not activate at all, because hypergraphs lose their anchor points back at the preprocessing stage: skeletonization and local reduction during their initial construction. This indicates the existence of a precision problem in the preprocessing mechanisms used in this model.

The second largest pattern, "9" → "4" (46 cases, 75.4% of class "9" errors), has a distinct nature. Attractors 4_1 and 9_2 coincide in the number of vertices (8), the number of edges (8), and the cyclomatic number (one loop), as well as the parameter values of the structural elements - meaning they are also indistinguishable in structural complexity. However, the nature of this indistinguishability is different. It is determined by the fact that when approximating the image contour with straight-line segments, the information contained in curved lines is lost. This specific information is critically important for separating the attractors of the digits "4" and "9".

**7. Discussion of Experimental Results and Prospects for Future Research.**

1. Experimental results confirm theoretical conclusions:

– For each class, the reduction operator converges to a fixed point in a finite number of steps. All 13 concepts stabilized; the number of iterations is finite and varies depending on the structural complexity of the digit. Attractor 7_1 (5 nodes) stabilized faster than 8_1 (11 nodes), which is consistent with the estimate of the number of steps via initial complexity in the proof of Theorem 1.

- The learning result is independent of the order in which originals are presented. The order of presentation of originals varied for each concept, but the resulting attractor (set of vertices and attribute intervals) remained unchanged. This is a direct empirical demonstration that the frequency function $f(e)$ depends only on the set of absorbed examples, not on the order of their absorption.
- The structure of the alphabet can emerge endogenously. Classes "1," "2," and "4" required more than one attractor for full coverage. This was not predetermined; the decision on the number of attractors per class was made iteratively based on the structural incompatibility of incoming examples. The subdivision into subclasses corresponds to actually observed stylistic writing variants of the digits.
- Decisions are fully interpretable at the local level. Each attractor is a structural object available for direct inspection (number of nodes, their types, coordinate intervals). Classification errors are explainable in terms of the model itself—no error is an inexplicable anomaly, unlike latent vectors in multi-dimensional space or their post-hoc explanations in surrogates like LIME and SHAP [36].
- A covering set consisting of a limited number of originals (3 to 11 examples per class) is sufficient for learning (convergence to an attractor) without optimization of an error functional.

Thus, the proposed method ensures the simultaneous fulfillment of three properties: single-pass learning without backpropagation, ultra-small sample size, and local interpretability at the model level itself.

2. Comparison of Modeling Results:

In this experiment, we did not strive to achieve SOTA results of 99.9% classification accuracy on MNIST. Our goal was to demonstrate PoC results for our theory. Given that we trained the model on an ultra-small sample, the obtained results should be compared with those of other Few-Shot Learning (FSL) models. The comparison results are presented in Table 5. It should be noted that this table represents a contextual comparison rather than a controlled benchmark.

Table 5. Comparison of simulation results.

| Model | Minimum number of unique samples | Augmentation | Epochs | Accuracy (%) | Source | Notes |
|---|---|---|---|---|---|---|
| **ISL (MNIST)** | **85 (3-11/ class)** | **Yes (30-110 per class – 784 total)** | **1** | **91.13** | **This work** | **No backpropagation, no hyperparameter tuning** |
| SVM (RBF) Reduced MNIST (RMNIST/5-10) | 50–100 (5-10/class) | - | 1 | 69–75 | Nielsen [37] | Requires hyperparameter tuning |
| MLP | 200 (20/class) | No/Yes | 100 | 53/61 | Overfitting problem [38] | Backpropagation, hyperparameter tuning, overfitting problem |
| CNN (FMNIST) | 100–200 (10-20/class) | Yes | 50 | 74–78 | Brigato et al. [39] | Backpropagation, hyperparameter tuning |
| Prototypical Networks (ProtoNet) | 5/class (5-way) miniImageNet | - | FSL (Metrics training) | 71 | Chen et al. [40], Snell et al. [41], | Pre-training in basic classes, using 5-way classification. |
| MAML (Model-Agnostic Meta-Learning) | 5/class (5-way) miniImageNet | - | FSL (Meta-learning) | 72 | Chen et al. [40] | Pre-training on task distribution, learning initialization parameters for fast adaptation. |
| CNN (RMNIST/5) | 50 (5/class) | Yes (500–1000) | 50 | 84.38 | Nielsen [37] | Backpropagation, dropout, hyperparameter tuning |

Although some of the compared results were obtained for datasets more complex than MNIST (such as miniImageNet or FMNIST), our model's accuracy of 91.13% on conditionally complete MNIST - compared to results like CNN (RMNIST/5) at 84.38%, MLP (MNIST) at 61%, and SVM (RMNIST/5-10) at 75% - serves as a compelling confirmation of its effectiveness from the perspective of the considered theory.

3. Discussion of Problematic Issues:

Most classification errors are driven by the following key challenges:

- Imperfections in graph edit distance calculation between hypergraphs, which underlies the WTA competition of attractors (i.e., the edit distance turns out to be non-monotonic with respect to the structural specificity of the attractor);
- Empirical choice of measured parameter types and their variation ranges;
- Significant reduction of detected/extracted primitives during the preprocessing stage;
- Substantial impoverishment of the parametric space during the reduction process, which calls for introducing additional measured parameters/metrics.

However, identifying these issues confirms that the error structure during attractor-based classification can be fully interpreted in terms of the model itself.

4. Prospects for Future Experimental Research:

We are continuing our work on the following tasks:

– Improving MNIST image classification quality by extracting (detecting) yet another type of primitive – curves - and expanding the initial parametric space;
– Associative recognition of MNIST images with incomplete contours, where individual structural elements belonging to the attractor (concept) have been lost;
– Scaling by the number of recognizable classes using examples of national alphabet symbols;
– Formation of scene attractors - the next level in the attractor hierarchy. We view scenes as complex/composite handwritten symbols represented in complex national alphabet datasets or the Omniglot dataset, such as Korean Hangul, Japanese Hiragana and Katakana, Khmer script, Tibetan script, Glagolitic, South Indian scripts, and others. We anticipate that at this stage we will be able to achieve SOTA results for FSL models featuring a large number of complex classes.

## IV. Neurobiological Hypotheses of Structural Projection and Neural Coding

In the experimental part of this work, we deliberately restricted the complexity of the problems under consideration to simple contour images and minimized the complexity of the hypergraphs by intentionally reducing the number of types of detected structural elements and their measurable parameters. As a result, we obtained relatively simple hypergraph structures. In addition, we modeled an active perceptual act using an attentional focus on the most informative structural elements, which represent the best candidates for the role of anchor points in the attractor structure. This allowed us to move from the general problem of subgraph isomorphism to the problem of isomorphism in constrained hypergraphs with low treewidth and vertices pre-labeled as candidate "anchors"; that is, we selected an input domain in which the hypergraph matching problem becomes polynomial-time solvable. In this case, this mitigates the NP-hardness of the problem but does not eliminate it when matching more complex hypergraphs representing realistic three-dimensional images, scenes, processes, and so on.

A number of approaches could be proposed to mitigate the problem of matching complex hypergraphs, for example:

- Hierarchical matching, in which large structural units are aligned first (connected components and global topology), followed, within each correspondence, by smaller units (local neighborhoods of anchors), and finally by parametric values. This mimics the way the human visual system operates from general to specific and structurally corresponds to the two-level reduction in ISL. Each level of the hierarchy drastically reduces the search space for the next level;

- Construction of learnable attention using Graph Neural Networks (GNNs) and gradient-based methods. In principle, this contradicts the ideology of ISL, but it may serve as an analytical tool that makes it possible to empirically investigate the properties of the operator $\mathcal{F}$ with the aim of its subsequent analytical formalization;
- The use of the parametric structure Σ***.** This direction is specific to ISL and has no direct analogue in the general theory of isomorphism. Anchors carry not only topological but also parametric information - the intervals Σ*. Matching can be reformulated as a joint optimization problem: simultaneously minimizing the structural distance $d_{\text{struct}}$ and the parametric distance $d_{\text{param}}$ (Equation 114). This transforms the problem from a purely combinatorial one into a mixed problem, in which the parametric part can be solved using continuous methods and can guide the search in the structural part. In practical terms, parametric compatibility provides additional filtering of candidates beyond structural constraints.

These approaches remain valid for computational implementations of ISL that do not claim biological plausibility, but they also do not eliminate the difficulty described above. Moreover, the main problem for the biological justification of our concept is that *hypergraph matching is not biologically plausible*, because the spatial structure of a dendritic tree is not topologically isomorphic to the stimulus. That is, a neuron does not possess a global map of the stimulus structure or a mechanism for its sequential comparison with a stored template; instead, it responds locally and in parallel, without exhaustively searching through possible correspondences.

To preserve biological plausibility, we assume that the solution to this problem may require changing the formulation of the problem itself - namely, abandoning hypergraph matching. To justify this conceptual reconsideration and formalize an alternative formulation, we propose four neurobiological hypotheses.

**1. Material Attractor.**

In the ISL model considered above, a hypergraph can be regarded as a mathematical/informational representation of a stimulus.

However, the dendritic tree of a neuron is not an informational but a physical/material object, i.e., an internal structure of a biological system. Let us assume that this structure can also be represented as a hypergraph at time $t$:

$$H_t = (V_t, E_t, \Theta_t) \quad (124),$$

where:

- $V_t$ denotes functional dendritic elements (synaptic structures);
- $E_t$ denotes hyperedges defining possible interactions (relationships);
- $\Theta_t$ denotes the parameters of these interactions.

In this case, the stimulus $\mathcal{G}_S$ (represented by a hypergraph in terms of the ISL model considered above) is not transformed into an internal hypergraph of the system. Instead, it is projected onto an already existing hypergraph $H_t$ - the structure of the neuron's dendritic tree, with functional relationships among synapses and synaptic structures (for example, dendritic spines), which form modal groups analogous to those in the ISL model.

Thus, the hypergraph $H_t$ is not a representation of an external stimulus but constitutes an internal “model of the world,” formed by the entire previous history of the system.

By the history of the system $\mathcal{H}_t$ - understood as the history of perception, life activity, or learning - we will, in a simplified form, mean the set of all previously perceived stimuli $\mathcal{G}_i$:

$$\mathcal{H}_t = \{\mathcal{G}_1, \mathcal{G}_2, \dots, \mathcal{G}_t\} \quad (125).$$

Then,

$$H_t = \Lambda(\mathcal{H}_t) \quad (126),$$

where $\Lambda$ is an operator of structural-parametric adaptation (plasticity).

History does not merely modify an individual parameter. Rather, it gradually shapes the architecture of the system's internal model itself - in this case, a *material structural attractor* realized on the hypergraph of the dendritic tree $H_t$ of a biological neuron.

**2. Structural Projection.**

Let us introduce a *structural projection operator* $\Pi$:

$$\Pi: \mathcal{G}_S \to H_t \quad (127).$$

This projection is not a mapping between vertices and does not define a topological isomorphism between hypergraphs. Rather, it is a structural projection determined by the structure of neuronal interactions. That is, we effectively assume that:

$$\mathcal{G}_S \ncong H_t \quad (128).$$

Thus, $\mathcal{G}_S$ and $H_t$ may have completely different topologies, while the projection induces similar reduction dynamics.

Then, if a stimulus $\mathcal{G}_S$ is presented,

$$\Pi: \mathcal{G}_S \times H_t \to \Delta H_t \quad (129),$$

where

$$\Delta H_t \subseteq H_t \quad (130)$$

is the activated part of the internal hypergraph.

Thus, the stimulus activates the internal structure not through spatial correspondence, but through the *correspondence of the dynamics of structural-parametric reduction*.

The operator $\Pi$ satisfies a consistency condition with $\varphi$:

$$\Pi(\mathcal{G}_S, H_t) \subseteq \varphi(H_t) \tag{131}$$

that is, the projection activates only structurally relevant elements of the material hypergraph.

### 3. Dynamic Equivalence.

***Definition 11.*** *Dynamic Equivalence.*

A stimulus $\mathcal{G}_S$ and an internal structure $\Delta H_t$ are called *dynamically equivalent* if there exist sequences of reductions $R_{\mathcal{G}}$ and $R_H$ such that

$$R_{\mathcal{G}}^n(\mathcal{G}_S) \to \mathcal{A}_{\mathcal{C}}^{(\mathcal{G})} \tag{132}$$

and

$$R_H^m\big(\Pi(\mathcal{G}_S, H_t)\big) \to \mathcal{A}_{\mathcal{C}}^{(H)} \tag{133},$$

where it is not topology that is preserved, but rather the structure of the convergence process.

The attractors themselves are therefore not required to be topologically identical; however, they must be equivalent with respect to their reduction dynamics.

That is, there must exist a mapping:

$$\Upsilon: \mathcal{A}_{\mathcal{C}}^{(\mathcal{G})} \to \mathcal{A}_{\mathcal{C}}^{(H)} \tag{134}$$

that preserves the relevant invariants.

In this case, we can define the relation:

$$\mathcal{A}_{\mathcal{C}}^{(\mathcal{G})} \sim_R \mathcal{A}_{\mathcal{C}}^{(H)} \qquad (135),$$

if both attractors induce the same class of stable structural invariants.

Thus, in the original ISL model, the attractor has the form $\mathcal{A}_{\mathcal{C}}^{(\mathcal{G})}$ and preserves the geometry of the symbol. For example, for the digit "8," it will still resemble the digit "8." In the dendritic model, however, a fundamentally different structure $\mathcal{A}_{\mathcal{C}}^{(H)}$ emerges. It may bear no resemblance whatsoever to the digit "8" and may be distributed throughout the dendritic tree in a complex manner.

However, if the condition $\mathcal{A}_{\mathcal{C}}^{(\mathcal{G})} \sim_R \mathcal{A}_{\mathcal{C}}^{(H)}$ is satisfied, both structures possess equivalent functional properties:

– they have equivalent invariant anchor cores: $E^{*\mathcal{G}} \sim V_t$;
– they have equivalent sets of parametric fibrations: $\sum^{*} \sim \Theta_t$;
– they have functionally equivalent relationships among anchor elements: $f(\mathcal{R}^{(p)}) \sim f(E_t)$ - the parametric hyperedges of the stimulus and the connections of the dendritic hypergraph may be completely different objects; nevertheless, they implement *the same constraints on the process of structural reduction*. This means that the parametric hyperedges of the stimulus and the functional connections of the internal hypergraph in the dendritic tree induce *the same class of admissible dynamics*;
– they converge to a fixed point in a finite number of reduction steps: $n \sim m$.

Thus, the issue is no longer one of structural isomorphism, but rather of the equivalence of dynamical systems.

This interpretation is consistent with neurobiological evidence. A neuron clearly does not store a geometrical copy of a face, a word, or a digit. Nevertheless, different implementations of a representation can lead to the same recognition, the same invariance, and the same behavior of the system.

## 4. Adaptation Operator and Initial Redundancy of the Material Hypergraph.

However, the general adaptation operator $\Lambda$ of the material hypergraph must have a more complex structure than the generalized reduction operator in the ISL model. It should consist of two components implementing two complementary mechanisms:

$$\Lambda = R \circ \Omega \quad (136),$$

where:

- $\Omega$ is the operator of structural-parametric expansion;
- $R$ is a generalized reduction operator analogous to that used in the ISL model.

The *structural-parametric expansion operator* $\Omega$ adds new elements to $H_t$ under the influence of a stimulus:

$$\Omega: H_t \times \Pi(\mathcal{G}_S, H_t) \to H_{t+1} \supseteq H_t \quad (137).$$

To implement such an adaptation operator, we may assume the existence of an initial redundancy in the structure of the material hypergraph - the dendritic tree.

***Assumption 1.*** *Initial Redundancy.*

The initial material hypergraph $H_0$ is redundant with respect to the set of attractors $\{\mathcal{A}_\mathcal{C}\}$ that the system is capable of forming:

$$C(H_0) \gg C(\mathcal{A}_\mathcal{C}) \forall \mathcal{C} \quad (138).$$

This means that the reduction operator $R$ can reveal structural attractors in $H_0$ without adding new elements.

It should be clarified that the initial structure $H_0$ contains an excess of potential degrees of freedom rather than hidden (pre-existing) attractors.

Two operating modes of the adaptation operator can therefore be distinguished.

***Mode 1.*** *Initial Learning ($H_0$ is redundant).*

As long as Assumption 1 holds, the expansion operator $\Omega$ is not required, since reduction of the redundant $H_0$ is sufficient:

$$H_{t+1} = R\big(H_t \oplus \Pi(\mathcal{G}_S, H_t)\big) \qquad (139),$$

where $\oplus$ denotes the accumulation operation in the ISL model.

The structural redundancy of $H_0$ provides the possibility of revealing attractors without adding new elements.

Mode 1 remains active as long as $H_t$ does not become a fixed point:

$$R\big(H_t \oplus \Pi(\mathcal{G}_S, H_t)\big) \neq H_t \qquad (140),$$

that is, as long as reduction continues to reveal a latent attractor within the redundant $H_0$.

It should be noted that a “fixed point” is a computational/mathematical idealization rather than a biological cessation of plasticity.

***Mode 2.*** *Adaptive Learning.*

When $C(H_t)$ approaches $C(\mathcal{A}_C)$ and structural redundancy has been exhausted, but there is a need to incorporate new or modified stimuli, the introduction of the structural-parametric expansion operator $\Omega$ becomes necessary:

$$H_{t+1} = R\left(\Omega\big(H_t, \Pi(\mathcal{G}_S, H_t)\big)\right) \qquad (141).$$

The *transition to Mode 2* occurs when two conditions are simultaneously satisfied:

$$R(H_t) = H_t \text{ and } \Pi(\mathcal{G}_S, H_t) \nsubseteq \mathcal{A}_{\mathcal{C}}^{(H)} \quad (142),$$

that is:

- if $H_t$ has reached a fixed point, then the reduction of $H_0$ has been exhausted. Neurobiologically, this corresponds to the end of the *critical period* - a limited developmental time window early in life during which the brain exhibits maximal synaptic plasticity and requires external stimuli for the proper formation of neural circuits - when *pruning* (the process of eliminating weak or unused synapses) has been completed;
- if a new stimulus introduces elements that are incompatible with the current attractor, reduction alone is insufficient. Neurobiologically, this corresponds to encountering a new or modified stimulus after the completion of the critical period, i.e., a situation requiring structural reorganization.

If $H_{t+1} = H_t$, that is $R(H_t) = H_t$ and $\Pi(\mathcal{G}_S, H_t) \subseteq \mathcal{A}_{\mathcal{C}}^{(H)}$, this defines stable recognition: the stimulus is fully compatible with the existing attractor, and no adaptation is required. Thus, a *reduction cycle* occurs.

Assumption 1 and the proposed operating modes of the adaptation operator have neurobiological support. For example, Huttenlocher, P. R. (1979) [42] showed that synaptic density in the human frontal cortex during the first years of life reaches approximately 150% of the adult level and subsequently decreases through pruning. Another study by the same author [43] confirmed that pruning in the visual cortex is influenced by visual experience.

In addition, study [44] shows that different cortical regions follow different trajectories of connectivity development. For example, the prefrontal cortex, visual cortex, and associative areas all exhibit distinct profiles of synaptogenesis - the process of forming new synapses between neurons for signal transmission - and synaptic pruning.

Such neurodynamics are qualitatively consistent with the assumption of an initial hypergraph $H_0$, which is gradually transformed under the influence of incoming stimuli and internal constraints.

***Assumption 2.*** *Local Adaptation and Global Invariants.*

1. Local Adaptation.

The operator $\Lambda$ acts locally: the adaptation of an element $e \in H_t$ is determined only by its participation in $\Delta H_t$, rather than by the global state of the neuron:

$$\Lambda(e) \neq 0 \Leftrightarrow e \in \Delta H_t \quad (143).$$

This assumption is analogous to the consideration of the process of identifying local anchors $E^*$ in the ISL model.

Recent *in vivo* studies of dendritic plasticity show that changes in the efficacy of individual synapses are determined by the local structure of activity within a dendritic compartment and may be relatively independent of the global output signal of the neuron (Wright et al., 2025 [45]). This result is qualitatively consistent with Assumption 2, according to which the structural adaptation of the internal hypergraph is local in nature and affects only the active fragment $\Delta H_t$, rather than the entire structure as a whole.

The results of Frey & Morris (1997) [46] are also consistent with Assumption 2 concerning the locality of structural adaptation. The mechanism of *synaptic tagging and capture* (STC) proposed by the authors demonstrates the existence of a biological process that makes it possible to identify a subset of active synaptic elements that subsequently become subject to long-term structural stabilization. In terms of our proposed model, such a subset can be regarded as a local active fragment of the hypergraph $\Delta H_t$, on which the structural reduction operator acts during the process of attractor formation.

Moreover, modern versions of the STC hypothesis consider memory formation as a problem of allocating limited plasticity resources among competing synaptic ensembles [47], which also supports our attractor-based interpretation.

2. Global Invariants.

We may also hypothesize that the role of the global invariants $\Phi_{\mathcal{C}}$ in the ISL model is performed in a neuron by the backpropagation of action potentials, or bAPs (*backpropagating action potentials*), which provide a *global signal* from the soma to dendritic compartments (action potentials propagate backward into the dendritic tree and activate voltage-dependent calcium channels) [48]. Other possible mechanisms underlying this process are also considered by Gerstner et al. (2018) [49].

In this case, the role of global invariants is not to preserve topology, but rather to establish (confirm the validity of) the global structure of the hypergraph attractor within the structure of the dendritic tree.

Then:

$$\Lambda + \text{bAP} \rightarrow \mathcal{A}_{\mathcal{C}}^{(H)} \quad (144).$$

That is, bAPs, together with other mechanisms, may serve as an instructive signal for the formation of the global structure of the attractor $\mathcal{A}_{\mathcal{C}}^{(H)}$. Through the mechanism of STDP (*Spike-Timing-Dependent Plasticity*), each neuronal firing event contributes an increment to the materialization of $\mathcal{A}_{\mathcal{C}}^{(H)}$ within the structure of $H_t$: the coincidence of local activation $\Delta H_t$ with a bAP determines which elements $v_i$are stabilized into the global topology of the attractor.

The global invariant $\Phi_{\mathcal{C}}$, however, is not transmitted by the bAP as pre-existing information. Rather, it emerges as the cumulative result of accumulated bAP events across a set of examples belonging to a class $\mathcal{C}$. Since the efficacy of bAPs decreases with distance from the soma, the formation of the global structure of the attractor likely requires additional hierarchical mechanisms for propagating plasticity among dendritic compartments.

In study [50], the authors investigated dendritic spine elimination during motor learning and concluded that the probability of synapse elimination is simultaneously determined by local factors (the state of the immediate dendritic environment) and global factors (the role of the particular neuron in behavior and learning). Thus, spine elimination is not a random process, which also supports our Assumption 2.

**5. Subjectivity of Attractors.**

Let there be two systems: $H_t^{(1)}$ and $H_t^{(2)}$.

Since their histories are different, $\mathcal{H}_t^{(1)} \neq \mathcal{H}_t^{(2)}$, it follows that $H_t^{(1)} \neq H_t^{(2)}$.

Consequently, $\Pi_1(\mathcal{G}_S) \neq \Pi_2(\mathcal{G}_S)$ and their attractors may also differ: $\mathcal{A}_{\mathcal{C}}^{(1)} \neq \mathcal{A}_{\mathcal{C}}^{(2)}$.

However, classification based on subjective "models of the world" remains the same:

$$Class\left(\mathcal{A}_{\mathcal{C}}^{(1)}\right) = Class\left(\mathcal{A}_{\mathcal{C}}^{(2)}\right) \quad (145).$$

Thus, subjectivity is a consequence of differences in internal "models of the world" resulting from differences in the material substrates $H_t^{(1)}$ and $H_t^{(2)}$, as well as from differences in the histories of attractor formation, rather than a consequence of noise or uncertainty.

Then:

$$H_{t+1} = \Lambda\big(H_t, \Pi(\mathcal{G}_S)\big) \quad (146).$$

That is, the history of stimulus perception does not create the dendrite itself but modifies the already existing hypergraph - the dendritic tree.

Thus, after the projection of a stimulus $\Pi(\mathcal{G}_S)$, during the process of structural-parametric adaptation $\Lambda$, the internal parameters of $H_t$ are modified. That is, $\Lambda$ acts not on the stimulus $\mathcal{G}_S$ itself, but on its projection - the internal "model of the world."

Consequently, the attractor in a given system is a stable configuration of this internal "model of the world":

$$\mathcal{A}_{\mathcal{C}} \subseteq H_t \tag{147}$$

Thus, we can formulate the following hypothesis.

**6. Hypothesis of Dynamic Equivalence of Attractors under Structural Projection.**

***Hypothesis 1.***

There exists a class of internal (material) hypergraphs $H$ for which the projection operator $\Pi$ preserves the dynamics of structural reduction in the sense of the relation $\sim_R$ (Definition 11).

For such systems:

1. The attractors $\mathcal{A}_{\mathcal{C}}^{(H)}$ of the material system and $\mathcal{A}_{\mathcal{C}}^{(\mathcal{G})}$ of the stimulus space are topologically different but dynamically equivalent: $\mathcal{A}_{\mathcal{C}}^{(\mathcal{G})} \sim_R \mathcal{A}_{\mathcal{C}}^{(H)}$.
2. Repeated presentation of stimuli belonging to class $\mathcal{C}$ leads to the materialization of an attractor in the architecture of $H_t$ through a reduction cycle consisting of two operating modes of the adaptation operator $\Lambda$.
3. The subjectivity of perception is a structural property of the system that follows from differences in material histories $\mathcal{H}_t$, rather than from uncertainty or noise.

Thus, ISL provides a mathematical model of structural self-organization in the space of hypergraphs. The hypothesis of dynamic equivalence of attractors under structural projection suggests that analogous principles may underlie the organization of dendritic structures in biological neurons, where the structural correspondence between a stimulus and its internal representation is not computed explicitly but emerges as a result of the dynamics of structural projection onto a material attractor.

To summarize our view of the solution to the hypergraph matching problem in ISL, we assume that, within the structural projection model, the problem of finding a structural correspondence between a stimulus and its internal representation is not solved through explicit hypergraph matching. Instead, the correspondence emerges as a result of the reduction dynamics of the activated fragment of the material hypergraph. Consequently, the computational complexity associated with hypergraph matching procedures is transferred from the algorithmic level to the level of the physical dynamics of the system.

## 7. Hypothesis of Neural Coding.

Three major paradigms of neural coding have emerged in neuroscience:

1. Rate coding: information is encoded by the firing rate (or the number of spikes within a time interval); the neuron operates as a "pulse counter."
2. Temporal coding: information is encoded by the precise timing of spikes, interspike intervals, phase relative to oscillations, and similar temporal variables; firing patterns carry meaning.
3. Population coding: information is distributed across an ensemble of neurons; meaning is contained in the activation pattern of the population (group), rather than in an individual neuron.

Despite substantial differences among these theories, most of them associate the representation of information with observable neural activity - firing rate, temporal structure, or ensemble dynamics.

We hypothesize that the "neural code" is determined not by the parameters of the output signal or by population activity, but by an attractor understood as a process of convergence of internal structural dynamic. The observable response of a neuron is merely a *projection of this dynamic onto the space of measurable output signals*.

The structure $H_t$ is the material substrate on which this dynamic unfolds, but it is not itself the "code." The scheme of such "coding" can therefore be represented as:

$$\underbrace{H_t}_{substrate} \overset{R^n}{\rightarrow} \underbrace{\mathcal{A}_{\mathcal{C}}^{(H)}}_{attractor=code} \overset{\mathrm{O}}{\rightarrow} \underbrace{output\ signal}_{code\ projection} \quad (148),$$

where O may be regarded as an operator for "reading out" the *current dynamics of the attractor*. In a simplified form, the current dynamical state of the attractor may be expressed (projected) through its complexity (Equation 60). However, as demonstrated by our ISL experiments on MNIST, such a linear projection - and even more complex dependencies used in the experiments - does not provide an adequate mapping ("readout") of the attractor, or its dynamics, into the output signal of a neuron. This is also consistent with the well-known fact that action potential generation in a biological neuron is a complex nonlinear process.

Thus, it can be assumed that the projection O of the current attractor into an output signal is not a "readout" of a fixed state but a dynamic process. Therefore, different parameters of the neuronal response (firing rate, burst length, and others) may reflect the convergence of the attractor toward a fixed point in different ways, and the choice of a particular parameter is an empirical question. For example, homeostatic plasticity may maintain the average firing rate, but not necessarily other parameters, such as burst duration, pattern regularity, or the temporal structure of the response.

It should be noted that, in the ISL model, the learning process is separated from the inference process, and the classification decision is not made through a direct comparison of the complexities of different simultaneously activated attractors (excited neurons). Instead, it results from a WTA (*Winner-Take-All*) competition procedure, in which the complexity of attractors is represented indirectly through the hypergraph edit distance $d$ (Equation 114). In this formulation, the complexity of hypergraph structures is also represented linearly, and we do not project the dynamics of attractor formation into an output signal, since we analyze only its final structure after the learning stage.

Returning to neurobiology, we hypothesize that a distributed ensemble of simultaneously activated neurons at the same level of the structural hierarchy (Theorem

3) is not itself the carrier of a concept. Rather, it constitutes a space of competing projections - a space of competing alternatives - whereas the concept itself is localized in the structural attractor of a single neuron (Theorems 1 and 2).

*The attractor of a neuron is projected into its response rather than constituting an exact encoding of a particular image. This projection is local, i.e., it has meaning only for the "selected" neurons within its neighborhood.*

That is, in practical terms, *a neuron "encodes" not the stimulus itself but the decision that has been reached (its internal justification).* This view is supported by some neurobiological studies [51]. We may therefore assume that any specialized competitive neural map necessarily contains a transition from a distributed space of alternatives to a localized carrier of the winning attractor.

A more radical assumption can be made if we adopt the definition of information proposed in Y. Parzhyn (2025) [2]. If information is understood not as a signal but as an internal structure of interrelated parameters forming the system's internal energy model, then such information is not transmitted through axons at all. Axons transmit only signal-events that initiate changes in the internal dynamics of the next neuron, whereas information emerges only as a result of the interaction of these events with the internal structural attractor of the dendritic tree.

Consequently, an axon does not carry a concept and does not explicitly encode it; rather, it merely initiates a reconfiguration of the dynamics of the subsequent attractor, while the meaning of the signal is determined not by the signal's own structure but by the internal organization of the postsynaptic neuron.

***Hypothesis 2.***

The observable response of a neuron does not encode a concept and does not transmit it to the next neuron. It represents a local projection of the internal attractor dynamics, interpreted only by those neurons whose dendritic structures have been formed through a shared process of development and learning. Therefore, the same activity pattern may have different functional meanings in different neurons and neural networks.

Within the framework of the definition of information proposed in *Architecture of Information* [2], information is localized neither in the parameters of neural activity nor in the transmitted signals as such, but in the internal structure of attractors that form a model of the "external world" of a given neuron.

An important consequence of this hypothesis is the multidimensional nature of the observable neuronal response. The very generation of an action potential may determine the qualitative outcome of recognition, whereas the parameters of the response may reflect quantitative characteristics of both the attractor itself and the process of convergence toward it. In particular, the neuronal response may contain:

- a relatively stable (invariant) component associated with the structural attractor and determining the fact of recognition itself;
- a dynamic component reflecting the influence of excitatory postsynaptic potentials (EPSPs) that are not part of the attractor structure and characterize the current context of stimulation. These EPSPs are not necessarily completely suppressed during dendritic processing and may contribute to the formation of the observable response without changing the recognition outcome;
- the type of excitation (according to ISL theory): direct Bottom-up excitation during inference; associative Bottom-up excitation during inference with an incomplete input signal vector; contextual Top-down excitation; and excitation during learning, involving Bottom-up + Top-down signals. The type of excitation may be reflected in the dynamics of the neuronal response. This is consistent with neurobiological observations. Several distinct firing modes have been described in pyramidal neurons, including tonic/regular spiking, bursting, high-frequency BAC-bursting associated with coincidence detection (*Burst-Attenuated / Backpropagation-Activated Calcium-spike firing*) [52], persistent firing, and others [53].

In this interpretation, the neuronal response is not a direct representation or "encoding" of a concept, but an observable projection of the internal dynamics formed by the interaction between the attractor and the current context.

It also follows that the actual mechanisms of neuronal competition may be substantially more complex than their simplified algorithmic representation in ISL theory. Competition is likely implemented through a set of interconnected processes, including lateral inhibition, local dendritic processing of signals, synaptic filtering, and other mechanisms operating at different levels of neural organization.

However, in our view, the problem of "neural coding" cannot be reduced exclusively to the analysis of the neuronal response itself. The response is only an intermediate component of a more complex process of information formation. To understand how the activity of presynaptic neurons is interpreted by postsynaptic neurons, it is necessary to consider the next level of organization, associated with the structure of the dendritic tree and the principles governing the formation of inter-neuronal connections.

It is precisely at this level that the spatial arrangement and clustering of synaptic contacts, electrotonic distance, temporal windows of coactivation, local NMDA spikes, dendritic plateaus, and other mechanisms become fundamental. These mechanisms determine how the same pattern of neuronal activity will be interpreted by different neurons. Consideration of this level leads to a hypothesis concerning the origin of stable structures of inter-neuronal projections and the formation of architectural constraints for the subsequent emergence of conceptual attractors.

Thus, in the proposed model, the informational meaning arises neither during signal generation nor during its transmission, but during its interpretation by the receiving neuron. If this is indeed the case, a fundamental question arises: *how is the architecture of the dendritic tree and inter-neuronal projections itself formed, thereby determining the future interpretation of incoming signals?*

## 8. Hypothesis of the Projection Attractor.

ISL theory assumes that the modeled neural network has local connections in the sensory system between receptors and primitive detectors - pre-detector neurons - as well as a fully connected architecture between pre-detector neurons and detector neurons for simple (contour) holistic images (concepts). This is technically justified but biologically implausible. In contrast, the proposed hypotheses assume the formation of a biologically plausible selection of inter-neuronal connections and relate their architecture to the structural projection operator Π. However, the formalization of this operator remains an unresolved issue.

Neurobiological data indicate that connections are formed through the coordinated influence of multiple factors: genetic, stimulus-dependent (external), and internal (physiological). This is difficult to formalize in detail. Nevertheless, we can propose a hypothesis for the formation of inter-neuronal connections (the projection operator) that is consistent with the ISL concept and has a neurobiological interpretation.

*8.1. Formal Problem Formulation.*

Let there be a set *M1* of primary primitive detectors with a deterministic retinotopic projection. Each neuron in *M1* detects a local primitive at a specific spatial position. In terms of the MNIST experiments conducted in this work, these include line segments, their orientation and length; structural points (endpoints, corners, intersections, junctions); and so on. The "address" of each neuron in *M1* in space is known.

There is also a set *M2* consisting of concept-detector neurons with an initial material dendritic structure $H_0$ containing an excess of potential degrees of freedom (Assumption 1).

The problem is to determine the principle (mechanism) governing the formation of interconnections between the outputs of neurons in *M1* and the dendritic structures of a neuron in *M2*, under the condition that these connections are formed neither according to a predefined map nor randomly. Formally:

$$\Pi: M1 \to M2 \quad (149).$$

We assume that the structural projection operator Π is not specified *a priori*, but emerges as an attractor of the connection-selection process from an overcomplete space of potential projections. This assumption is fully consistent with the proposed ISL concept.

Let us assume that the morphology of the dendritic tree $H_0$ of each neuron in *M2* is given and is not itself the subject of this hypothesis.

*8.2. Space of Potential Projections.*

Let us assume that a set of external and internal factors defines a set of potentially admissible connections

$$\Pi_0 = \{(d_i, n_j, v)\} \quad (150),$$

where a connection between detector $d_i$ and dendritic segment $v$ of neuron $n_j$ is considered physically admissible.

Mathematically, $\Pi_0$ defines the complete combinatorial space of anatomically possible synaptic contacts and is considered an overcomplete set of possible projections.

From a neurobiological perspective, $\Pi_0$ may manifest itself as the combinatorial redundancy of synaptogenesis during the early stages of neuronal development. However, construction and formalization of $\Pi_0$ require additional information concerning axonal projections, growth mechanisms, molecular markers, connection-addressing rules, and so on.

Therefore, the *existence of a space of potentially admissible connections* $\Pi_0$ *is taken as an initial condition*, while its specific structure is determined by factors whose formalization lies beyond the scope of this work.

*8.3. Hypothesis of the Existence of a Projection Attractor*.

***Hypothesis 3.***

We hypothesize that the structural projection operator $\Pi$ emerges as an attractor of the reduction process $F_{dev}$ acting on the space of potential projections $\Pi_0$, which is determined by the composition of two interconnected processes represented by the operators of statistical coactivation $Co$ and dendritic selection $DS$.

Formally:

$$F_{dev} = DS \circ Co \quad (151),$$

where the strict sequence of composition is a simplification. Then:

$$\Pi = Fix(F_{dev}) \quad (152),$$

where we will refer to $F_{dev}$ as the *relationship development operator*.

Then there exists a dynamic process:

$$\Pi(t+1) = F_{dev}\big(\Pi(t)\big) \quad (153).$$

Thus, in the general case, we show that there are not only structural attractors of the type $\mathcal{A}_{\mathcal{C}}$, but also attractors of projection dynamics $\Pi$.

Let us consider this hypothesis.

Suppose that the process of forming inter-neuronal projections can be described by the relationship development operator $F_{dev} = DS(H_0) \circ Co$, acting on the space of potential projections $\Pi_0$ that emerges during the development of the nervous system.

If this process possesses a stable fixed point $\Pi = Fix(F_{dev})$, then this fixed point is interpreted as a *projection attractor* that determines a stable organization of inter-neuronal connections.

The projection attractor defines the representational space within which conceptual attractors $\mathcal{A}_{\mathcal{C}} = Fix(R)$ subsequently emerge.

*Coactivation (Statistical Selection) Operator.*

Among anatomically available candidates from *M1*, connections are preferentially formed between neurons that are systematically coactivated. Coactivation is determined by the statistics of the incoming stimulus stream during the critical period: neurons detecting primitives that frequently occur together in natural scenes are activated in a coordinated manner. Through the STDP mechanism, this coordinated firing strengthens the synaptic connections between them [54]. Thus, the statistics of structures in natural stimuli are not random; they contain structural regularities. It is possible that these regularities, together with the genetic program, determine the topology of the connections being formed.

Formally, we can define the coactivation operator as:

$$Co(i,k) = \frac{1}{T}\int_0^T \phi_i\,(t)\phi_k(t)\,dt \qquad (154),$$

where $\phi_i(t)$ and $\phi_k(t)$ denote the activities of the neurons. The more frequently two neurons are activated together, the higher the value of $Co(i, k)$. This operator is essentially a formalization of the Hebbian principle, although in this form it should be regarded primarily as illustrative.

*Dendritic Selection Operator.*

The dendritic tree $H_0$ does not passively accept all strengthened synapses; its geometry creates preferential zones of integration. Dendritic branches with particular lengths and branching patterns have characteristic temporal integration windows: synapses located close to one another on the same branch are integrated more effectively than distant synapses. This may mean that a neuron in *M2* "prefers" synaptic inputs whose spatiotemporal pattern corresponds to its dendritic geometry. Thus, the initial geometry of the dendritic tree $H_0$ acts as a kind of structural template that resonates with certain coactivation patterns from *M1*. Connections whose patterns resonate with the geometry are stabilized, whereas the others are eliminated through pruning.

Let us consider this reasoning in more detail.

A key element of the projection attractor hypothesis is the dendritic selection operator $DS(H_0)$, which links the morphological organization of a neuron to the process of forming stable inter-neuronal projections.

Modern neurobiological research shows that dendrites are not passive conductors of electrical signals. The efficacy of synaptic input integration is determined not only by its amplitude but also by its spatial location on the dendritic tree, the temporal coordination of activation, the electrotonic distance from the soma, and local dendritic nonlinearities [55]. In addition, studies of synaptic clustering [56] show that functionally related inputs tend to stabilize on the same dendritic branches, forming local computational modules.

These findings make it possible to consider the dendritic tree not as a passive carrier of connections, but as an active computational structure possessing its own selectivity with respect to input patterns.

Let Equation (124) define the initial morphology of the dendritic tree of neuron $n_j$. For each dendritic segment $v_j$, let us introduce a characteristic temporal integration window, simplified as a function of signal propagation delay (although the actual window depends on many factors, including membrane time constants, conductances, geometry, ion channels, and others):

$$\theta(\mathrm{v}) = \lambda \frac{L\left(v_j \ \rightarrow soma\right)}{v} \qquad (155),$$

where $L\left(v_j \rightarrow soma\right)$ is the electrotonic distance from the segment to the soma, "v" is the effective propagation velocity, and $\lambda$ is a scaling coefficient.

The quantity $\theta(\mathrm{v})$ defines the characteristic temporal scale within which input signals are most effectively integrated by a given dendritic segment.

Consider two primitive detectors $d_i$ and $d_k$, for which the characteristic temporal shift between their activations in the input stream is $\Delta t_{ik}$. We assume that this pattern is in a state of *structural resonance* with segment $v_j$ of the dendritic tree of neuron *M2* if the following condition is satisfied:

$$|\Delta t_{ik} - \theta(v)| \leq \varepsilon \qquad (156),$$

where $\varepsilon$ defines the admissible deviation from the optimal temporal window.

To quantitatively evaluate the degree of such correspondence, we introduce the dendritic selection operator

$$DS(i, k, v) = \exp\left(-\frac{\left(\Delta t_{ik} - \theta(v)\right)^2}{2\varepsilon^2}\right) \qquad (157).$$

Values of $DS(i, k, v)$ close to one correspond to high compatibility between the temporal structure of the input pattern and the characteristics of the dendritic segment. Values close to zero correspond to weak compatibility and a low probability of long-term stabilization of the corresponding connections.

It is important to emphasize that the presented form of the operator $DS(H_0)$ is not a direct consequence of existing neurobiological models. Experimental data support the existence of dendritic selectivity, the dependence of integration on morphology, and the presence of local mechanisms for strengthening coordinated inputs [56]. However, the specific form of the function $DS(H_0)$ represents a hypothesis proposed within the ISL framework as a possible simplified mathematical model of this process.

In this interpretation, the dendritic tree acts not only as a physical substrate for signal processing but also as a structural functional for selecting input patterns. Different dendritic tree morphologies implement different selection operators $DS(H_0)$. Consequently, even under identical input-stream statistics, different neural systems may form different stable connection structures.

Thus, the projection attractor is determined by the joint action of environmental statistics and the internal morphology of the system:

$$F_{dev} = DS(H_0) \circ Co, \qquad F_{dev}: \Pi_0 \to \Pi_0 \qquad (158),$$

where the operator $Co$ reflects the statistical regularities of the input stream (154), while the operator $DS(H_0)$ represents the structural constraints imposed by the morphology of the dendritic tree.

Within this hypothesis, the projection attractor represents a stable organization of inter-neuronal connections determined by the joint action of statistical regularities of the environment and morphological constraints of the system.

Then the structural projection operator (152) takes the form:

$$\Pi = Fix(DS(H_0) \circ Co), \qquad DS(H_0) \circ Co: \Pi_0 \to \Pi_0 \qquad (159).$$

The complete causal chain of the proposed hypotheses can now be represented as follows:

$$H_0 \to DS(H_0) \to \Pi = Fix(DS(H_0) \circ Co) \to \mathcal{A}_{\mathcal{C}} = Fix(R) \qquad (160),$$

where Hypothesis 1 describes the transformation

$$H_0 \xrightarrow{R} \mathcal{A}_{\mathcal{C}} \qquad (161),$$

where $\mathcal{A}_{\mathcal{C}}$ is the attractor of the space of structures, while Hypothesis 3 describes the transformation

$$\Pi_0 \overset{DS(H_0)\circ Co}{\rightarrow} \Pi \quad (162),$$

where Π is the attractor of the space of projections.

Hypothesis 3 directly implies *projection subjectivity*, which can be formalized as:

$$H_0^{(1)} \neq H_0^{(2)} \Rightarrow DS(H_0^{(1)}) \neq DS(H_0^{(2)}) \Rightarrow \Pi_1 \neq \Pi_2 \quad (163),$$

that is, two systems may receive the same stream of stimuli but form different representational spaces. This consequence demonstrates yet another level of subjectivity of attractors.

Thus, unlike the structural attractor $\mathcal{A}_{\mathcal{C}}$, whose existence is proven within the formal theory of ISL, the existence of the projection attractor Π is considered a hypothesis in the present work.

**9. Hypothesis of Counter (Cross-System) Learning.**

In the ISL theory proposed in Section II and its experimental validation in Section III, we essentially consider supervised learning, since we define the classes and their labels ourselves. However, the problem of supervised learning is fundamental from the standpoint of biological plausibility. The proposed hypothesis is aimed at addressing this problem. We show that cross-system coactivation can serve as a trigger both for the process of forming inter-neuronal connections and for the process of attractor formation without supervision in the classical sense of this ML paradigm.

Hypothesis 3 postulates a general mechanism ($Co$, $DS$, $F_{dev} = DS \circ Co$) and states that it converges to a stable structure interpreted as a projection attractor, but it leaves unresolved the question: "Why does activity arise in a potential postsynaptic neuron if the connection with the presynaptic neuron has not yet been formed?". Hypothesis 3 does not consider the role of Top-down signals in the formation of neuronal connections and the structural attractors themselves.

We propose the following Hypothesis 4, which provides a general mechanism for addressing three related problems:

- the possibility of coactivation between a presynaptic primitive and a postsynaptic element whose connection has not yet been formed;
- formation of their interconnection based on this coactivation;
- formation of a structural attractor based on coactivation between different systems.

Thus, Hypothesis 4 is intended to explain *the origin of the projection attractor itself* and to show how cross-system coactivation of already formed attractors belonging to different functional systems of the organism can serve as an internal teaching signal, eliminating the need for an external teacher and predefined class labels in the sense of classical machine-learning paradigms. At the same time, global biological reinforcement mechanisms may play a modulatory role without determining the content of the attractor being formed.

*9.1. The Term "Counter Learning".*

***Definition 12.*** *Counter Learning.*

*Counter (cross-system) learning* is defined as the process of forming a new structural attractor in one functional system under the influence of the joint activity of local sensory representations and an already formed structural attractor in another functional system.

The term *Counter learning* was introduced in the work of Y. Parzhin et al. (2020, *Neurocomputing*) [1], where the same counter-learning scenario is described that is formalized in greater detail in the proposed Hypothesis 4. This term refers to a mechanism different from the earlier term *Counterpropagation Networks* (Hecht-Nielsen, 1987), which has a different architecture and purpose, as well as from the independently proposed later term *Counter-Current Learning* (Kao & Hariharan, NeurIPS, 2024), which considers gradient-based credit assignment through paired local losses and therefore fundamentally differs from the learning mechanism discussed in Hypothesis 4.

*9.2. Hypothesis of Counter Learning.*

***Hypothesis 4.***

The formation of new structural attractors occurs through cross-system coactivation of previously formed and newly forming conceptual representations belonging to different functional systems of the brain. An already formed structural attractor in one system recruits a free concept-neuron in another system, initiating structural plasticity, the formation of directed inter-neuronal connections, and subsequent structural reduction, which leads to the emergence of a new structural attractor.

Let system $k$ have a set of primitives $M_1^{(k)}$ and a concept-neuron from $M_2^{(k)}$ with a material substrate $H^{(k)}$. Consider two systems: *Base* and *New*.

Suppose that in the *Base* system there is an established stable attractor that is currently activated/excited. This attractor may, for example, be a neuron-detector in the auditory perception system activated by auditory sensors.

$$\mathcal{A}_{Base} = R^{T_1}(H_0^{Base}) \quad (164),$$

where $R^{T_1}$ is the reduction operator and $T_1$ is the number of convergence steps of the material substrate $H_0^{Base}$ toward the attractor $\mathcal{A}_{Base}$.

Thus, the *Base* system does not necessarily have to be a system of another sensory modality. It may be any other system with a stable, coactive representation: the speech system, motor system, spatial-attention system, interoceptive (physiological) system, proprioceptive system, emotional system, or another system.

The *New* system is the system in which we consider the process of forming connections between a pre-detector neuron (primitive detector) $M_1^{New}$ and a detector neuron (concept-neuron) $M_2^{New}$ under the control of $\mathcal{A}_{Base}$. This system may, for example, be the visual system.

According to Assumption 1, neuron $M_2^{New}$ possesses a redundant initial substrate $H_0^{New}$. Thus, $\mathcal{A}_{Base}$ and $H_0^{New}$ are not parts of the same hypergraph.

We also consider two projection spaces. The space $\Pi^{New}$ is the same space as $\Pi_0$ in Hypothesis 3, but for the *New* system: the formalized set of potentially admissible connections between primitives $M_1^{New}$ and segments of the dendritic tree. It is precisely in this space that the substantive part of the hypothesis unfolds.

$$\Pi^{New} \subseteq M_1^{New} \times H_0^{New} \quad (165).$$

The element

$$p = (d_i, v) \in \Pi^{new} \quad (166)$$

denotes a potentially admissible connection between the axon of pre-detector neuron $d_i \in M_1^{new}$ and segment $v \in H_0^{new}$ of the dendritic tree of neuron $M_2^{New}$.

The space $\Xi$ defines the connection between the *Base* and *New* systems. The existence of this connection is assumed as an initial condition and will not be considered in detail in this hypothesis.

$$\Xi \subseteq \{\mathcal{A}_{Base}\} \times M_2^{new,free} \quad (167),$$

where $M_2^{new,free}$ is a free, not yet activated neuron of the *New* system.

Let us consider the three main phases of counter learning.

***Phase 1.*** *Recruitment of a free neuron* $M_2^{new,free}$.

The activated attractor $\mathcal{A}_{Base}$ generates a top-down signal $\psi_{base}(t)$, which propagates through the existing inter-system pathways.

For each free neuron $n_i \in M_2^{new,free}$, define the recruitment function

$$\xi(n_i, t) = \mathbf{1}\left[\psi_{base}(t) > \tau_{n_i}\right] \quad (168),$$

where:

- $\mathbf{1}[\cdot]$ is the indicator function;
- $\tau_{n_i}$ is the minimum level of top-down activation required to bring a neuron into the recruitment state.

When $\xi(n_i, t) = 1$, the neuron is considered selected for the formation of a new concept.

The existence of the pool of free neurons $M_2^{new,free}$ with an initial, statistically redundant dendritic structure (Assumption 1) does not require prior activity: the basic architecture of synaptic contacts can emerge even in the complete absence of neurotransmitter transmission (Verhage et al., 2000 [57]). However, this local mechanism of contact formation provides only a sufficient condition, i.e., the mere possibility of local, activity-independent contact formation does not determine which free neuron will be recruited by a given $\mathcal{A}_{Base}$ to form the "correct" structure of inter-neuronal contacts. In our view, this specificity may be provided by the capture procedure proposed in Parzhin (2020) [1].

At this stage, the top-down signal does not form new connections; it only determines which free neuron will become the carrier of the new attractor.

***Phase 2.*** *Coactivation and transition to the state of structural plasticity.*

The recruited neuron receives two independent sources of excitation.

The first is the top-down signal $\psi_{base}(t)$. The second is the bottom-up activity of a set of pre-detectors $M_1^{new}$,induced by the stimulus $\mathcal{G}_S$..

Let the activity of pre-detector $d_i$ be denoted by $\phi_i(t)$. Then the set of simultaneously active pre-detectors is defined as

$$M_1^{act}(t) = \left\{d_i \in M_1^{new} \mid \phi_i(t) > \tau_{up}\right\} \quad (169),$$

where $\tau_{up}$ is the minimum level of bottom-up activation of a neuron.

In a simplified form, we assume that prolonged joint coactivation of $\psi_{base}(t)$ and $\{\phi_i(t)\}$ brings the recruited neuron into a *state of structural plasticity*:

$$Ә(n, t) = 1 \quad (170),$$

where:

$$Ә(n, t) = \mathbf{1}\left[\int_{t-T}^{t} \psi_{base}(\tau) \cdot \Phi^{act}(\tau)\, d\tau > \Theta_{Ә}\right] \quad (171),$$

where:

$$\Phi^{act}(\tau) = \sum_{d_i \in M_1^{act}(\tau)} \phi_i(\tau) \quad (172).$$

Here:

- $T$ is the coactivation window;
- $\Theta_{Ә}$ is the threshold for transition to the state of structural plasticity.

Thus, Hypothesis 4 assumes that the state Ә = 1 serves as a *trigger for known cellular mechanisms of structural plasticity*, such as activity-dependent synaptotropic stabilization of contacts (Vaughn, 1989; Niell, Meyer & Smith, 2004), including processes of growth, remodeling, and stabilization of dendritic and axonal contacts [58, 59].

The specific molecular mechanism is not specified in this work and is not the subject of the present investigation.

***Phase 3.*** *Formation of directed connections.*

After the transition to the state Ә = 1, each potential contact $p$ (166) is evaluated according to the condition of local spatiotemporal coactivation.

Let us define the set of stabilized projections:

$$\Delta H_{expand} = \{p = (d_i, v) \in \Pi^{new} | \ni(n,t) = 1, \phi_i(t) > \tau_{up}, Co(p) > \Theta_C \} \quad (173),$$

where:

- $Co(p)$ is the local measure of coactivation at the presumed contact point (Hypothesis 3); here, the $Co$ operator is applied in an extended sense: $v$ acts as a local analogue of the neuron-partner in coactivation;
- $\Theta_C$ is the stabilization threshold for a new connection.

The resulting set of stabilized connections expands the initial architecture:

$$H_0^{new'} = H_0^{new} \oplus \Delta H_{expand} \quad (174).$$

After that, structural reduction is performed:

$$\mathcal{A}_{new} = R^{T_2}\left(H_0^{new'}\right) \quad (175),$$

which leads to the formation of a new structural attractor.

Thus, we assume that cross-system coactivation determines not the geometry of future inter-neuronal connections, but rather the selection of the concept-neuron and the initiation of structural plasticity processes, whereas the specific architecture of new connections is formed through intra-system coactivation and local mechanisms of contact growth and stabilization.

*9.3. The Role of Top-Down Signals.*

It is assumed that the same top-down signal performs *three different functions* at different stages of the attractor life cycle.

*1. Connection Formation Phase.*

The top-down signal performs the *recruitment of a free concept-neuron*, determining the carrier of the future structural attractor and bringing it into a state of structural plasticity.

*2. Learning Phase.*

After the formation of the primary connections, the same top-down signal performs the function of a *learning trigger*, ensuring the joint coactivation of the recruited neuron and active pre-detector neurons required for the formation of a structural attractor through reduction.

*3. Functional Phase.*

After learning has been completed, the top-down signal performs the function of *associative activation*: in the absence or incompleteness of bottom-up inputs, it can directly activate the formed attractor, initiating processes of completion and associative recognition.

*9.4. Consequences of Hypothesis 4.*

***Assumption 3.*** *On the Existence of Innate Attractors.*

There exists a finite set of evolutionarily formed basic attractor states (or an *attractor basis*) possessing the properties of fixed points of the reduction operator and providing the initial conditions for the ontogenetic formation of subsequent conceptual attractors. Their origin belongs to the phylogenetic time scale and may be interpreted within the framework of the *Baldwin effect* as the result of selection and genetic assimilation of innate predispositions facilitating the formation of repeatedly occurring adaptive structures [60].

***Consequence 5.*** *On the Role of the Number of Cross-System Coactivation Events.*

Hypothesis 4 implies that *the potential for the formation of structural attractors is determined not only by the number of independent functional systems, but also by the number of possible acts of their cross-system coactivation that implement the mechanism of counter learning.*

**10. Conclusions of the Section.**

The fourth part of this work proposes a system of neurobiological hypotheses whose purpose is not to construct a new model of the neuron, but rather to explore a possible biological realization of the mathematical principles of ISL theory. Unlike the formal part of the work, in which the existence of a structural attractor has been rigorously proven, all propositions considered below are hypothetical in nature and require independent experimental verification.

The central idea of the proposed approach (Hypothesis 1) is the *transfer of the concept of a structural attractor from the abstract space of hypergraphs to the space of intracellular dynamics*. Within this hypothesis, it is assumed (Hypothesis 2) that a *concept is represented not by the observable response of a neuron, but by a stable internal structure of dendritic dynamics, whereas the action potential represents only its local projection*.

This assumption leads to a fundamental reinterpretation of the problem of neural coding. The observable neuronal response is proposed to be considered not as a direct carrier of information, but as an intermediate projection of an internal attractor state. The informational meaning itself is not transmitted along axons in a ready-made form; rather, it arises only through the interaction of incoming signals with the internal organization of the receiving neuron. This interpretation is directly consistent with the definition of information proposed in *Architecture of Information* [2], according to which information does not exist independently of the perceiving system but is formed through the process of its internal organization.

The present work also proposes to consider the observable response of a neuron as a multidimensional projection of attractor dynamics. Different parameters of the action potential - such as firing rate, temporal structure, burst activity, latency, and other characteristics - may reflect different aspects of the internal attractor and the process of convergence toward it, rather than representing competing mechanisms of neural coding.

This provides a unified interpretative framework for different existing theories of neural activity coding.

The next level of the proposed model is Hypothesis 3, concerning the *architectural (projection) attractor*. It is assumed that the stable organization of interneuronal connections itself emerges as the result of an earlier attractor process of development. In this case, the space of potentially possible dendritic projections is reduced to a stable architecture within which conceptual attractors of individual neurons subsequently emerge. Thus, two levels of self-organization are distinguished: the formation of the architecture of interneuronal interactions and the formation of conceptual attractors within an already established architecture.

A special role is assigned to Hypothesis 4, which introduces the *mechanism of counter (cross-system) learning*. In this approach to learning, an already established structural attractor in one functional system initiates the formation of a new structural attractor in another system through top-down projections. Unlike classical models of learning involving an external teacher or reinforcement signal, the proposed model assigns the role of a learning signal to sustained cross-system coactivation of the organism's functional systems.

Hypothesis 4 does not contradict contemporary views concerning the local nature of synaptic plasticity and structural growth. It is assumed that top-down activation does not directly form new connections but instead brings a neuron into a state of structural plasticity by activating known cellular programs of axonal and dendritic growth. The formation of specific synaptic contacts is then mediated by local activity-dependent and activity-independent developmental mechanisms.

Thus, in the proposed model, top-down signals acquire multiple functional roles. At different stages of concept formation and functioning, they may act as:

- triggers for the formation of new interneuronal connections;
- triggers for learning and the formation of a structural attractor;

- mechanisms of associative activation of an already established attractor in the absence or insufficiency of bottom-up signals.

Within the proposed model, the potential for the formation of structural attractors is determined not only by the number of independent functional systems but also by the number of possible acts of their cross-system coactivation that implement the mechanism of counter learning. This makes it possible to interpret the development of cognitive complexity in an organism as a consequence of the expansion of the space of interactions among sensory, motor, physiological, and other functional systems, rather than exclusively as the result of accumulating statistical experience within each individual system.

Hypothesis 4 is conceptual in nature and requires further experimental verification. Nevertheless, it proposes a possible mechanism for the emergence of new concepts without an external teacher by linking structural plasticity, dendritic growth, cross-system coactivation, and the formation of structural attractors within a unified framework.

The proposed hypotheses make it possible to reinterpret a number of known neurobiological phenomena, including local dendritic plasticity, synaptic tagging and capture, spatial clustering of synapses, lateral neuronal competition, and the selectivity of neuronal detectors. At the same time, we do not claim that existing experimental neurobiological findings directly confirm the proposed model; rather, they are considered independent observations that may be interpreted within the framework of the proposed theory.

The main result of this part of the work is therefore not a proof of the validity of the proposed hypotheses, but the formulation of a system of experimentally testable consequences of the mathematical theory of ISL. In this way, the neurobiological part moves the theory of structural attractors from the domain of an abstract mathematical model into the domain of potentially falsifiable neurobiological propositions and identifies directions for future research linking mathematical modeling, computational experiments, and modern experimental neuroscience.

# V. Conclusion

The present work has attempted to develop a coherent formal model of Invariant Structural Learning (ISL), integrating a mathematical description of the process of concept formation, an experimental demonstration of the feasibility of the proposed approach, and its possible neurobiological interpretation.

In the theoretical part of the work, a mathematical framework for ISL was developed based on the representation of objects as hypergraphs and their successive structural reduction. It was shown that the reduction process can be considered a dynamical system converging to a stable structural attractor, which is interpreted as the concept of a class. The proposed definitions and theorems formalize the fundamental properties of the model and make it possible to consider learning as a process of forming stable structural representations rather than as the statistical optimization of model parameters.

The experimental part of the work demonstrates the fundamental feasibility of the proposed approach. Despite the substantially smaller amount of training data and the absence of error backpropagation, the obtained results demonstrate the possibility of forming stable structural representations from a limited number of positive examples. Thus, the experiments do not establish the superiority of the proposed model over modern machine-learning architectures for a particular classification task. Rather, they demonstrate the possibility of an alternative, non-statistical mechanism of learning based on the self-organization of structural and parametric invariants.

The proposed neurobiological interpretation is not presented as a proven description of brain function. Instead, it formulates a system of interrelated hypotheses that makes it possible to interpret known neurobiological observations within a unified conceptual framework. In this interpretation, information is localized not in the parameters of neuronal responses but in the structure of an internal attractor, while the observable response represents only a local projection of this internal dynamic. The introduced concepts of the projection attractor and the cross-system formation of new attractors make it possible to propose a potential mechanism for the emergence of concepts without an external teacher, linking learning processes to the self-organization of interacting functional systems of the organism.

At the same time, the present work deliberately limits the scope of the questions considered. It does not investigate the mathematical mechanisms of local parametric reduction used in constructing hypergraph representations of objects, methods for the automatic identification of candidate anchor points, quantitative models of dendritic dynamics, or the mechanisms underlying the formation of innate attractors and their phylogenetic evolution. These questions require independent investigation and may be regarded as natural directions for the further development of the proposed theory.

The proposed theory does not claim to provide a definitive description of the mechanisms of intelligence. However, in the authors' view, it establishes a formal framework that makes it possible to consider the processes of concept formation, learning, and the structural organization of information within a unified mathematical and biological perspective. Testing the validity of this framework will require further theoretical and experimental investigation.

If the proposed principles prove to be valid, learning may be reconsidered as a process of the self-organization of structural attractors that constitute an internal model of the perceived world.

***Links.***

A description of the implementation of the ISL software model, ComAN, and an analysis of the experimental results are publicly available on GitHub at: https://github.com/NaturalAGI-Lab/NaturalAGI.

***Declaration on the Use of AI.***

During the preparation of this manuscript, the authors used generative large language models (LLMs), including Claude Sonnet 5, GPT-5.6 Luna, Gemini 3.6 Flash, and earlier versions of these models, for the following purposes:

1. Literature search and preliminary analysis;
2. Review and editing of the manuscript, including verification of logical consistency and mathematical notation;
3. Generation of Figure 1 based on clearly specified instructions provided by the authors;
4. Translation of the manuscript into English.

The authors personally verified all data and references generated with the assistance of LLMs and assume full responsibility for the overall integrity and content of this publication.